\documentclass[10pt]{article} % For LaTeX2e

\usepackage[accepted]{tmlr}
\usepackage{amsmath,amsfonts,bm}

\newcommand{\newterm}[1]{{\bf #1}}

\def\eqref#1{equation~\ref{#1}}
\def\1{\bm{1}}

\def\vz{{\bm{z}}}

\def\mX{{\bm{X}}}

\DeclareMathAlphabet{\mathsfit}{\encodingdefault}{\sfdefault}{m}{sl}
\SetMathAlphabet{\mathsfit}{bold}{\encodingdefault}{\sfdefault}{bx}{n}

\newcommand{\E}{\mathbb{E}}

\newcommand{\R}{\mathbb{R}}

\newcommand{\cI}{\mathcal{I}}

\usepackage{hyperref}
\usepackage{url}
\usepackage[dvipsnames]{xcolor}
\usepackage{framed}
\usepackage{enumitem}
\usepackage{algorithm}
\usepackage{algpseudocode}
\usepackage{subcaption}
\usepackage[most]{tcolorbox}
\usepackage{tikz}
\usepackage{graphicx}
\usepackage{lastpage}
\usepackage{makecell}
\usepackage{booktabs}
\usepackage{amssymb}
\tcbset{
    simbox/.style={
        enhanced,
        colback=white,
        colframe=black,
        boxrule=0.5pt,
        arc=1pt,
        left=2pt,
        right=2pt,
        top=2pt,
        bottom=2pt,
        boxsep=1pt,
        width=\linewidth,
        valign=top
    }
}
\newcommand{\panelimg}[2]{%
    \par\noindent
    \begin{tikzpicture}[baseline=(img.south)]
        \node[anchor=south west, inner sep=0pt] (img) at (0,0)
            {\includegraphics[width=\linewidth]{#2}};
        \node[
            anchor=north west,
            fill=white,
            fill opacity=0.85,
            text opacity=1,
            inner sep=1.5pt,
            font=\bfseries\scriptsize
        ] at ([xshift=2pt,yshift=10pt]img.north west) {#1};
    \end{tikzpicture}%
    \par
}

\newcommand{\rowlabel}[1]{%
  \begin{minipage}[t]{0.035\textwidth}
    \vspace{0pt}
    \small \textbf{#1.}
  \end{minipage}%
}
\newcommand{\rowcontent}[1]{%
  \begin{minipage}[t]{0.955\textwidth}
    \vspace{0pt}
    #1
  \end{minipage}%
}

\newcommand{\labeledrowcontent}[4]{%
\rowcontent{%
    \begin{tikzpicture}
        \node[anchor=north west, inner sep=0pt] (img) at (0,0)
            {\hspace*{#4}\includegraphics[width=#2]{#3}};
        \node[
            anchor=north west,
            fill=white,
            fill opacity=0.85,
            text opacity=1,
            inner sep=2pt,
            font=\bfseries
        ] at ([xshift=2pt,yshift=-2pt]img.north west) {#1};
    \end{tikzpicture}%
}%
}

\title{Cluster LOCO: Feature Importance for Interpreting Clusters}

\author{\name Claire He \email cmh2277@columbia.edu \\
       \addr Department of Statistics\\
       Columbia University \\
       New York, NY 10027, USA 
      \AND
      \name Genevera Allen \email genevera.allen@columbia.edu \\
      \addr Department of Statistics\\
       Columbia University \\
       New York, NY 10027, USA }

\def\month{11}  % Insert correct month for camera-ready version
\def\year{2026} % Insert correct year for camera-ready version
\def\openreview{\url{https://openreview.net/forum?id=QfKlFWtoAW}} % Insert correct link to OpenReview for camera-ready version

\begin{document}

\maketitle

\begin{abstract}
Clustering is widely used for exploratory analysis and scientific discovery, driving insights from market segmentation to biological data analysis, but its outputs can be difficult to interpret, audit, and reproduce as modern datasets become increasingly large and complex. Reliable use of clustering requires understanding which features drive the discovered structure, yet feature-level explanations for clustering remain scarce compared with methods in supervised learning. Furthermore, existing clustering feature importance scores are often tied to specific algorithms and data assumptions. To address these challenges, we propose Cluster LOCO (Leave-One-Covariate-Out), a family of model-agnostic feature importance scores for clustering. Cluster LOCO is built on feature occlusion and clustering generalizability, defined as whether cluster labels learned on one subset of the data can be accurately predicted on held-out samples. For any chosen clustering algorithm, Cluster LOCO quantifies a feature’s importance by measuring how much its removal degrades generalizability. We first introduce Cluster LOCO-Split, which relies on data splitting, and then extend it to Cluster LOCO-MP, a minipatch ensemble-based version designed for large-scale data. Across synthetic simulations and an application to cell-type discovery in single-cell transcriptomics, we show that Cluster LOCO more reliably recovers informative features than existing clustering feature importance methods. 
\end{abstract}

\section{Introduction}
A fundamental task in unsupervised learning, clustering is used across disciplines ranging from the social sciences, astrophysics to biology \citep{handcock_model-based_2007, materne_structure_1978, xu_clustering_2010} to draw insights from data by forming groups or partitions. Yet clustering is not defined by one canonical objective. As emphasized by~\citet{luxburg_clustering_2012}, clustering has several use cases: it may be used for preprocessing, to organize, compress, or denoise data; for exploration, to reveal unknown structure and generate hypotheses; or for confirmation, to validate hypothesized groupings or support scientific discoveries. Because clustering intervenes upstream in the data science life-cycle \citep{yu_veridical_2020}, its outputs influence downstream analysis, modeling, and scientific interpretation. It is therefore consequential that clustering-driven conclusions be reliable, trustworthy, and reproducible.

\noindent At the same time, \textit{reliability} is difficult to assess for clustering solutions because of the underlying assumptions and choices made by clustering models: for example, K-means requires specifying the number of clusters, and the algorithm will return exactly that many groups whether or not such structure is meaningful in the data \citep{allen_interpretable_2023}. This challenge is amplified in modern datasets, where complex nonlinear patterns, interactions, and high dimensionality can limit the effectiveness of classical methods \citep{kriegel_clustering_2009}. In response, practitioners increasingly rely on deep clustering models  \citep{min_survey_2018, li_attention-based_2023} or heavily feature-engineered workflows \citep{ding_k_2004, jolliffe_principal_2016, jin_influential_2016, stuart_comprehensive_2019, wolf_scanpy_2018}. While these approaches can improve the detection of complex structure (e.g. nonlinearity) or domain-dependent specificity (e.g. zero-inflated data), they can also make the resulting clusters harder to understand and audit. For instance, in genomics, clustering has enabled meaningful discoveries of cell types and markers \citep{villani_single-cell_2017} while, on the other hand, computational studies centered on clustering have raised persistent concerns about reproducibility \citep{gibson_perspectives_2022}. For clustering to support rigorous discovery, we therefore need tools that clarify why a clustering solution arises. In particular, feature-level explanations can help identify which features the clustering solution relies on and promote \textit{trust} \citep{gan_are_2025}.

\noindent We address this need by bringing the perspective of interpretable machine learning (IML) to clustering, establishing a useful notion of feature importance for clustering solutions. In many clustering applications, features are themselves meaningful and interpretable: they are the genes in transcriptomic data, words in text data, behavioral or measured attributes in the social sciences and often the object of downstream analysis. For example, in single-cell genomics, practitioners commonly interpret clusters by identifying “marker genes” i.e. genes that differ across the discovered clustering groups \citep{villani_single-cell_2017}, through differential expression analysis, where genes are tested across clusters \citep{kiselev_challenges_2019}. While useful for annotation, this workflow can raise post-selection inference, or “double dipping” concerns because the same data are used both to define the clusters and to assess the features that distinguish them, a practice known to inflate the false discovery rate of significant genes \citep{zhang_valid_2019, denadel_knockoff_2024, song_clusterde_2023}. Moreover, features that differ across clusters are not necessarily the features that produced the clustering solution. We therefore focus on feature importance for clustering, which asks directly how much each feature contributes to the clustering structure itself.

\noindent This IML perspective is well established in supervised learning, where feature importance methods are widely used to explain model predictions \citep{molnar_interpretable_2018}. In clustering, however, feature-level interpretability remains comparatively underdeveloped. Existing work has largely followed two directions: intrinsically interpretable clustering algorithms, built on interpretable supervised methods such as decision trees with a modified objective for clustering \citep{hu_interpretable_2024}, and post-hoc explanations tailored to specific algorithms, mainly K-means \citep{napoles_feature_2024, kauffmann_clustering_2024}. While useful, these approaches can be difficult to scale to large datasets, might make particular data assumptions or constrain the practitioner to specific clustering models. On the other hand, feature selection in clustering enables handling large scale data and has been well studied with models leveraging sparsity via regularization \citep{witten_framework_2010, wang_sparse_2018}, filters or wrapper methods \citep{xing_cliff_2001, dash_feature_2002, roth_feature_2003,  alelyani_feature_2014}. While related, feature selection and feature importance answer different questions. Feature selection asks which features should be used to construct a clustering solution, often by optimizing a sparsity or clustering quality criterion. In contrast, feature importance asks, after a clustering solution has been obtained, which features contributed most to it. We instead seek a post-hoc interpretation of a chosen clustering model, separating the explanation of a clustering solution from the decisions used to produce it.

\noindent To formulate such a post-hoc interpretation, we turn to feature importance methods from supervised interpretable machine learning. In supervised learning, feature-level interpretability includes a wide range of model-specific and model-agnostic techniques. For model-agnostic techniques, we find three main types of feature importance metrics: feature permutation introduced via model-class reliance \citep{fisher_all_2019, breiman_random_2001} inspects the model’s "reliance" to each feature via its error; Shapley values \citep{shapley_value_1953, lundberg_unified_2017} and extensions \citep{sundararajan_many_2020, mase_cohort_2021, verdinelli_feature_2023}, a popular metric based in game-theoretical axioms which distributes feature value across features fairly; and feature ablation or occlusion which explains the change in prediction when a feature is removed \citep{lei_distribution-free_2017,rinaldo_comment_2019, verdinelli_feature_2023}.

\noindent In this paper, we propose a model-agnostic feature importance framework for clustering based on feature occlusion. Our approach is motivated by the Leave-One-Covariate-Out framework \citep{lei_distribution-free_2017}. We assess feature importance through the change in clustering generalizability under feature occlusion: a feature is important if its removal makes the clustering solution harder to generalize on new observations. Our definition is post-hoc as it is applied after a clustering method has already been chosen, and is model-agnostic, therefore flexible to complex clustering workflows. Our contributions are as follows: first, we introduce Cluster LOCO, a family of model-agnostic feature importance metrics for clustering based on feature occlusion via Cluster-LOCO-Split. Second, we develop Cluster LOCO-MP, a scalable minipatch-based extension designed for high-dimensional data. Third, we validate our approach in simulated settings, including low-dimensional examples with complex nonlinear structure and high-dimensional regimes, and compare it against existing feature importance methods. Lastly, we demonstrate our framework in a single-cell transcriptomics application where important features for clustering are biologically meaningful.

\section{Feature importance scores for clustering via LOCO}

\subsection{Background for Cluster LOCO}
Before introducing our Cluster LOCO family of metrics, we review two existing notions that inspired our metric: generalizability and LOCO.

\newterm{Generalizability} in clustering, also sometimes coined as \newterm{predictability}, was first proposed in the context of model selection by \citet{lange_stability-based_2004} and \citet{tibshirani_cluster_2005} as an alternative to stability-based validation techniques relying on sampling or bootstrapping non disjoint subsets of data \citep{ben-hur_stability_2001, levine_resampling_2001}. Instead, the stability of a cluster solution is captured by a transfer predictor to measure how generalizable the cluster solution on one disjoint subset is to another. We apply this idea to the ML paradigm of data splitting, where a portion of the data is carved out as a training set, another held-out for calibration i.e. $\mX = \mX_{tr} \sqcup \mX_{cal}$. Then the measure of generalizability of the clustering solution is defined as the error (or dissimilarity) between the clustering solution on the calibration set ($\vz_{cal}$ cluster labels for $\mX_{cal}$) with its transferred labels via a transfer classifier $\hat f_{tr}$ trained on the training data ($\mX_{tr}$ with cluster labels $\vz_{tr}$).

\begin{equation}
\text{generalizability error} = \text{Error}\big(\vz_{cal}, \hat{f}_{tr}(\mX_{cal}; (\mX_{tr}, \vz_{tr})\big)
\label{eq:gen}
\end{equation}

\noindent This quantifies the \textit{generalizability} of clustering solutions: a small value for the error (or dissimilarity) between clustering solutions and predictions on the calibration set means the clustering solution learned from the train set is very generalizable to the held-out calibration set. Because the training and calibration sets are clustered independently, their cluster labels are arbitrary up to permutation. For label-dependent error measures, we therefore align the calibration labels $\vz_{cal}$ with the predictions $\hat f_{tr}(\mX_{cal})$ of the full-feature transfer classifier using Hungarian matching \citep{kuhn_hungarian_1955}, which places the calibration clustering in the label space learned from the training clustering. This alignment is computed once using the full-feature classifier and held fixed when evaluating all reduced classifiers $\hat f_{tr}^{-j}$. Examples of such error measures include the mean squared error (used observation-wise in \cite{tibshirani_cluster_2005}'s \textit{prediction strength} index), or the Hamming distance (which corresponds to the misclassification risk of the calibration set) in \cite{lange_stability-based_2004}. However, for clustering which involves multi-class labels, a more popular similarity metric for clustering label comparison is the adjusted-Rand index (ARI) which is permutation invariant. Alternatively, we also use the multi-class hinge loss as point-wise error measure for Cluster LOCO when using a soft classifier. These scores empirically yield similar normalized feature importance scores (see Figure~\ref{fig:supp_fig7} for a comparison of error measures on a simple example).

\newterm{Leave-One-Covariate-Out} (LOCO) on the other hand is an extensively studied quantity borrowed from supervised learning \citep{lei_distribution-free_2017, verdinelli_feature_2023, gan_model-agnostic_2023, little_iloco_2025} which quantifies the change in prediction error when removing a feature. In its LOCO-Split form, the LOCO error for feature $j$ given test point $(X, y)$ is obtained by the difference in prediction error on the test set in the absence of feature $j$ (i.e. $\hat f^{-j}_{tr}$ fit on without-$j$ training data $(\mX_{tr, -j}, \textbf{y}_{tr})$) and the prediction error on the test set given the full feature set for training (i.e. $\hat f_{tr}$ trained on $(\mX_{tr}, \textbf{y}_{tr})$) as shown in equation~\ref{eq:LOCO}.
 
\begin{equation}
\text{LOCO}_j(X,y) = \text{Error}\big(y, \hat{f}^{-j}_{tr}(X_{-j}; (\mX_{tr,- j}, \mathbf{y}_{tr})\big) - \text{Error}\big(y, \hat{f}_{tr}(X; (\mX_{tr}, \mathbf{y}_{tr})\big)
\label{eq:LOCO}
\end{equation}

Essentially, LOCO quantifies how much the model performance drops when retraining the model after excluding feature $j$ to determine whether it was an important feature: if the performance degrades, then feature $j$ is important whereas if the performance remains unchanged, feature $j$ isn't important.

\subsection{Cluster LOCO-Split: a generalizability feature importance score}

The LOCO objective lends itself naturally to a clustering extension: one may remove a feature, re-cluster the data, measure the resulting change in cluster solution using internal validity indices or stability scores for example. However, popular validity indices that would measure the effect of reclustering after removal such as the silhouette score \citep{rousseeuw_silhouettes_1987,karanikola_investigating_2021} often rely on geometric assumptions (i.e. favoring compact or well-separated clusters). On the other hand, most stability criteria \citep{ben-hur_stability_2001,luxburg_clustering_2009} measure the robustness of a clustering solution under resampling. A LOCO score based on such criteria would therefore primarily capture the sensitivity of the clustering algorithm to feature removal, making its interpretation as feature importance less direct.  

We instead use generalizability as a meaningful quantity for feature importance via LOCO: providing a model-agnostic feature importance -- \textit{agnostic} in the usual IML sense, applicable to any clustering algorithm the practitioner may use. We give an illustrative example of both silhouette-based LOCO and stability-based LOCO scores' limitations in Section 3 compared to our generalizability-based Cluster LOCO-Split score.

\newterm{Cluster LOCO-Split.} Let $\mX\in\mathbb{R}^{N\times M}$ be split into equal-sized training and calibration sets, $\mX_{tr}$ and $\mX_{cal}$, and let $C_\theta$ denote a clustering algorithm. We cluster the two subsets separately to obtain $\vz_{tr}=C_\theta(\mX_{tr})$ and $\vz_{cal}=C_\theta(\mX_{cal})$. Holding these cluster assignments fixed, we train a generalizability classifier $\hat f_{tr}$ on $(\mX_{tr},\vz_{tr})$ and its reduced counterpart $\hat f_{tr}^{-j}$ on $(\mX_{tr,-j},\vz_{tr})$. We then define Cluster LOCO-Split for feature $j$ as
\begin{equation}
\hat{\Delta}_j(\mX) = \underbrace{\text{Error}\big(\vz_{cal}, \hat{f}^{-j}_{tr}(\mX_{cal, -j})\big)}_{\text{generalizability error without } j} - \underbrace{\text{Error}\big(\vz_{cal}, \hat{f}_{tr}(\mX_{cal})}_{\text{generalizability error}} \big)
\label{eq:Cluster_LOCO_Split}
\end{equation}

Essentially, Cluster LOCO-Split quantifies how feature removal affects the transfer of cluster assignments from the training to the calibration set. If removing feature $j$ increases the transfer error, then it contributes to the generalizability of the learned cluster assignments across training and held-out observations. Algorithm~\ref{alg:split} summarizes the procedure. Specifically, Cluster LOCO-Split estimates the following population target, defined given a fixed training dataset $\mX_{tr}$ of $n$ observations following population distribution $P_{X^n}$, a clustering algorithm $C$, and fitted transfer classifiers $\hat f_{tr}$, $\hat f^{-j}_{tr}$, 

$$\Delta_j(\mathbf{X}_{tr}) = \E_{X \sim P_{X^n}}\big[\text{Error}\big(C(X), \hat f_{tr}^{-j}(X_{-j})\big) -\text{Error}\big(C(X), \hat f_{tr}(X)\big)\big|  \mX_{tr}, \hat f_{tr}, \hat f_{tr}^{-j}\big].$$

\noindent This target measures the expected change in transfer prediction error on an independent sample from the same population as the training data given a clustering algorithm (illustrated on a synthetic example in Figure~\ref{fig:supp_fig8a}).\medskip

\noindent Similar to the silhouette-based or stability-based LOCO score ideas, our generalizability-based score could incorporate feature removal during clustering. However, we find that by fixing the cluster assignments learned on all features, we can isolate the effect of generalizing the assignments. We also explored, in preliminary experiments, an alternative score
in which feature $j$ is removed before clustering both the training and calibration sets. This changes both the target partitions and the information available to the transfer classifiers, making it harder to distinguish changes in predictability from changes in the clustering structure itself. In preliminary synthetic experiments, we also found that this formulation yielded less stable feature importance scores.

\begin{algorithm}[htbp]
    \caption{Cluster LOCO-Split}
    \label{alg:split}
    \begin{algorithmic}[1]
    \Require Unlabeled $\mX \in \R^{N \times M}$ 
    \State Split data into {\color{Blue} training} and {\color{ForestGreen} calibration sets}:  $\mX_{tr} = \{X_i\}_{i \in \cI_{tr}}$ and $\mX_{cal} = \{X_i\}_{i \in \cI_{cal}}$.
    \State Cluster each split: 
    \begin{enumerate}[label=(\alph*),noitemsep, topsep=0pt]
    \item Cluster the {\color{Blue} training data}: $\{X_{i}\}_{i \in \cI_{tr}}$ to obtain cluster labels $\{\vz_i\}_{i \in \cI_{tr}}$.
    \item Cluster the {\color{ForestGreen} calibration data}: $\{X_i\}_{i \in \cI_{cal}}$ to obtain cluster labels $\{\vz_i\}_{i \in \cI_{cal}}$. 
    \end{enumerate}
    \State Fit generalizability classifiers and align generalizability if needed
    \begin{enumerate}[label=(\alph*),noitemsep, topsep=0pt]
        \item Fit $\hat{f}$ to the {\color{Blue} cluster-labeled training data} $\{(X_i, \vz_i)\}_{i \in \cI_{tr}}$.
        \item Fit $\hat{f}^{-j}$ to {\color{Blue} cluster-labeled without $j$ training data} $\{(X_{i, -j}, \vz_i)\}_{i \in \cI_{tr}}$.
    \end{enumerate}
    \State Compute \& return Cluster LOCO-Split $$\hat \Delta_j^{\text{split}} := \text{Error}\big(\vz_{cal}, \hat{f}^{-j}_{tr}(\mX_{cal, -j}| (\mX_{tr, -j}, \vz_{tr})) \big) - \text{Error} \big(\vz_{cal}, \hat f_{tr}(\mX_{cal}|(\mX_{tr}, \vz_{tr}))\big)$$
    \Ensure $\{\hat \Delta_j^{\text{split}} \}_{j=1}^M$
\end{algorithmic}
\end{algorithm}

\subsection{Scaling Cluster LOCO: a fast procedure for high-dimensional data}

\noindent Cluster LOCO-Split inherits several limitations from data splitting and feature occlusion. First, splitting the data reduces the sample size available both for fitting the clustering model and for estimating the generalizability error. This can reduce accuracy and make the score sensitive to the specific train-calibration split. The issue is especially pronounced when clusters are unbalanced: if either split contains few observations from a cluster, the resulting generalizability estimate can become unstable. Second, because Cluster LOCO-Split removes one feature at a time, correlated features can receive small importance scores: removing one may have little effect because the other remains in the active feature set. This can make it difficult to identify features that provide similar predictive information. 
Multi-fold splitting can mitigate some of the instability caused by data splitting, but it increases the computational cost substantially and may be impractical for large datasets as one would have to refit $F \times (M+1)$ models, where $F$ the number of folds. In this section, we introduce Cluster LOCO-MP, a minipatch-based extension designed to address these limitations. By leveraging ensembles of small random subsets of observations and features, Cluster LOCO-MP provides an ensemble-based Cluster LOCO score easy to parallelize and more stable when identifying important features, improving scalability in large datasets. 

\subsubsection{Cluster LOCO-MP: Cluster LOCO with minipatch ensembles}
To scale up our Cluster LOCO feature importance score for large-scale data, we introduce minipatches. Minipatches are tiny subsamples of the data that enable extremely parallel model fitting via ensembling. Instead of traditional subsampling observations into batches \citep{breiman_random_2001, louppe_understanding_2015}, minipatches require simultaneous subsampling of $n << N$ observations and $m << M$ features, and yield a natural structure of the data in in- and out-of-patch features and in- and out-of-patch observations. Minipatches have been shown to have computational advantages in the supervised learning setting \citep{yao_feature_2021, toghani_mp-boost_2021}, benefit from implicit ridge-like regularization useful in correlated settings \citep{yao_minipatch_2021} and have been used successfully for clustering in consensus clustering \citep{gan_fast_2022}. Drawing on these properties of minipatches, Cluster LOCO-MP is computationally advantageous for large datasets, operating under a prespecified computational budget determined by the number of minipatches, while feature subsampling can improve the identification of correlated signal features. More specifically, in the supervised setting, \citet{gan_model-agnostic_2023} propose a minipatch ensemble strategy to approximate the effect of feature occlusion. Building on this idea, Cluster LOCO-MP follows the same principle as Cluster LOCO-Split: a feature is important when withholding it worsens the prediction of cluster labels. Rather than predicting on a single held-out set, however, Cluster LOCO-MP aggregates predictions across many minipatches, yielding more stable importance estimates than a single data split (Figure~\ref{fig:supp_fig6}). While the two scores are closely related, the population target of Cluster LOCO-MP differs from that of Cluster LOCO-Split and therefore has a slightly different interpretation: the clustering being evaluated is now a consensus clustering constructed across minipatches, and generalizability is measured through an ensemble minipatch predictor rather than transfer classifiers fitted on a fixed training split. \medskip

%By drawing on those properties of minipatches, our Cluster LOCO-MP score is \textcolor{blue}{computationally advantageous with large datasets - operating under a prespecified computational budget determined by the number of minipatches, and can improve the identification of correlated signal features through feature subsampling. More specifically, in the supervised setting, \cite{gan_model-agnostic_2023} proposes a minipatch ensemble strategy to approximate the effect of feature occlusion. Building on this, Cluster LOCO-MP follows the same principle as Cluster LOCO-Split: a feature is important when withholding it worsens the prediction of cluster labels. But instead of predicting on a held-out set, Cluster LOCO-MP uses aggregated predictions across many minipatches to yield more stable importance estimates than with a single data split (Figure~\ref{fig:supp_fig6}). However, while related, the population target of this new score differs from Cluster LOCO-Split and should be interpreted slightly differently: the clustering evaluated is now a consensus-clustering stitched across many minipatches and the generalizability is an ensemble minipatch predictor rather than a transfer classifier trained on a fixed training split.} \medskip

 \noindent For fixed minipatch sizes $n$ and $m$ and number of minipatches $B$, let $\mathcal U$ denote the randomness in minipatch selection, clustering, and classifier training. We define the finite-sample Cluster LOCO-MP population target as
 
\begin{equation}
\Delta_{j}^{\mathrm{MP}}:= \mathbb E_{\mX \sim P_X^N,\,\mathcal U} \left[ \operatorname{Error}\!\left(  \vz^*, \hat H^{-j}(\mX_{-j}) \right) - \operatorname{Error}\!\left( \vz^*, \hat H(\mX) \right) \right],
\end{equation}

\noindent where the consensus labels $z^*$ and both ensembles are constructed from the same dataset and minipatch realizations. This target measures the expected change in prediction error when feature $j$ is excluded from the contributing
minipatches, compared to the minipatch consensus clustering solution (illustrated on synthetic data in Figure~\ref{fig:supp_fig8b}). Similar to Cluster LOCO-Split, this population quantity is defined for a given base classifier and clustering algorithm, but the interpretation is different: from a score for fixed training and fitted generalizability classifiers, we move on to interpret a minipatch-ensemble, consensus-clustering-based quantity. This is similar to the discrepancy between LOCO-Split interpretation and LOCO-MP \citep{gan_model-agnostic_2023}. 

\noindent Algorithm~\ref{alg:mp} presents the simplified full Cluster LOCO-MP procedure. The detailed algorithm outlining aggregation and label alignment strategies for hard and soft labels can be found in the additional materials (Algorithms \ref{alg:mp-soft} and \ref{alg:mp-hard}). % The algorithm follows the same principle as Cluster LOCO-Split, but replaces a single train-calibration split with an ensemble of minipatches in a leave-one-out framework exploiting the natural in and out-of-patch structure. 
Given a base cluster algorithm and classification algorithm, we specify in advance a large number of minipatches $B$. Then our algorithm can be summarized in four steps: first, performing minipatch clustering and generalizability training; second and third, obtaining the ensemble LOO and LOCO-LOO predictions; and finally computing the scores. Since we cluster each minipatch and then train a generalizability classifier to predict the resulting cluster labels, they are arbitrary across minipatches and require to be aligned to a common labeling. We outline in our algorithm an alignment step using a reference clustering but when computing such a reference clustering is costly, one can alternatively use the pairwise overlap across minipatches to get approximate alignment. The aligned predictors are then aggregated out-of-training: the LOO ensemble predictor $\hat H$ averages soft-predictions (or uses max-vote with hard-predictions) over minipatches that exclude the target observation, whereas the LOCO-LOO predictor $\hat H^{-j}$ further restricts this average to minipatches that also exclude feature $j$. The Cluster LOCO-MP score is then defined as the change in prediction error between the ensemble $\hat H^{-j}$ and $\hat H$.

\begin{algorithm}[htbp]
    \caption{Cluster LOCO-MP}
    \label{alg:mp}
    \begin{algorithmic}[1]
    \Require Unlabeled $\mX_1,\ldots,\mX_N\in\R^M$; minipatch sizes $(n,m)$ with $n<N$, $m<M$; number of minipatches $B$; error measure.

    \State Perform minipatch consensus clustering and training:
    \begin{enumerate}[label=(\alph*),noitemsep,topsep=0pt]
        \item Randomly sample $B$ minipatches of size $n\times m$ indexed by $(I_b,F_b)_{b=1}^B$.
        \item Obtain a reference clustering $\vz_0=C_\theta(\mX)$.
        \item Cluster each minipatch to obtain $\tilde{\vz}_b=C_\theta(\mX_{I_b,F_b})$. Align $\tilde{\vz}_b$ with $\vz_{0,I_b}$ via  Hungarian matching to obtain $\vz_b$, with entry $z_{b,i}$ for observation $i\in I_b$.
        \item Calculate consensus labels:
        \[z_i^*=\arg\max_{k\in[K]} \sum_{b:\,i\in I_b}\mathbf{1}\{z_{b,i}=k\}.\]
        Break ties using the smallest class index and leave $z_i^*$ undefined when no votes are available.
        \item Train each generalizability classifier $\hat f_b$ on $(\mX_{I_b,F_b},\vz_b)$.
    \end{enumerate}

    \State Construct the Cluster LOO ensemble prediction:
    \[\hat H_i:= \operatorname{Aggregate} \bigl(\hat f_b(\mX_{i,F_b}):b\in\mathcal B_i\bigr), \qquad \mathcal B_i=\{b:i\notin I_b\}. \]
    \State Construct the Cluster LOCO-LOO ensemble prediction: 
    \[\hat H_i^{-j}:= \operatorname{Aggregate} \bigl(\hat f_b(\mX_{i,F_b}):b\in\mathcal B_{i,-j}\bigr), \qquad \mathcal B_{i,-j}=\{b:i\notin I_b,\ j\notin F_b\}.\]
    Here, "$\operatorname{Aggregate}$" is the appropriate aggregation method for hard or soft classifiers  (Algorithms~\ref{alg:mp-soft} and~\ref{alg:mp-hard}).

    \State Compute Cluster LOCO-MP scores for $j=1,\ldots,M$:
    \[\hat\Delta_j:= \operatorname{Error}(\vz^*,\hat H^{-j}) - \operatorname{Error}(\vz^*,\hat H),\]

    \Ensure $\{\hat\Delta_j\}_{j=1}^M$.
    \end{algorithmic}
\end{algorithm}

\noindent The computational advantages of our method come primarily from the minipatch construction. Each minipatch contains only a small subset of observations and features, reducing the cost of fitting computationally expensive clustering procedures on large data. Moreover, because minipatches are generated and fitted independently, the training step is naturally parallelizable. For example, spectral clustering has time complexity $\mathcal{O}(N^3)$, which becomes $\mathcal{O}(Bn^3)$ under minipatch clustering, where $n \ll N$ is the number of observations in each minipatch and $B$ is the number of minipatches specified given the users' available computational resources. The ensemble scores can also be computed in parallel across features, making the implementation modular and adaptable to the available computational resources. Beyond computational efficiency, feature subsampling in the minipatch ensemble helps evaluate the contributions of correlated features. Because different minipatches contain different subsets of features, some include one feature of a correlated group but not another, while others exclude the group altogether. The resulting ensemble prediction error therefore reflects how much predictive information is lost when a feature is unavailable across the ensemble, even when it belongs to a set of correlated features - in the correlated designs considered, Cluster LOCO-MP recovers signal features more consistently than Cluster LOCO-Split in Figure~\ref{fig:supp_fig6}.

\subsubsection{Cluster LOCO-RAMPART}
In high-dimensional applications, practitioners are often primarily interested in the top-\(k\) ranked features driving a clustering solution rather than the full set of important features. This is particularly relevant in applications such as genomics, where many features may be irrelevant or redundant (e.g. house-keeping genes) and interpretation typically focuses on a small set of candidate markers. In such cases, computing Cluster LOCO-MP scores for every feature can be both computationally expensive and statistically inefficient. We therefore combine Cluster LOCO with the RAMPART framework of \citet{chen_top-k_2025}, a data-adaptive procedure for top-\(k\) feature ranking. In RAMPART Cluster LOCO, Cluster LOCO-MP is used as the round-level feature scoring oracle: at round $t$, Cluster LOCO-MP produces importance scores $\{\hat\Delta_j^{(t)}:j\in S_t\}$ on the active feature set $S_t$. Features are then ranked from these scores, the bottom half is discarded, and the procedure repeats until only the top-$k$ candidates remain. This yields an adaptive feature importance procedure: preliminary Cluster LOCO-MP estimates are used to screen and prioritize candidate features, and additional computation is concentrated on the most promising features retained through multiple rounds of fixed-batch sequential halving. \citet{chen_top-k_2025} showed that while theoretically the order of the number of minipatches required with this procedure is roughly the same as for the minipatch estimates, empirically, in their work and also in our sparse high-dimensional simulations in section 3, this adaptive procedure improves the identification of important features while reducing computational cost substantially.  The underlying interpretation remains unchanged: features are important when their removal increases the generalizability error of the clustering solution. The full Cluster LOCO-RAMPART algorithm is provided in the Appendix (Algorithm~\ref{alg:rampart}).

\subsection{Extensions and practicalities}

\subsubsection{Cluster-specific LOCO}
In many applications, cluster-level feature interpretability is more informative than global feature interpretability: for instance, in genomics, this corresponds to understanding which genes are important for a particular cluster that might identify with a specific cell type. We defined earlier the Cluster LOCO importance metric in the most general sense, compatible with dissimilarity metrics as well as point-wise error scores. In the case of point-wise error scores like multi-class hinge loss, we can extend the family of Cluster LOCO global importance scores to cluster-level importance scores. Cluster LOCO-Split admits a natural cluster-specific interpretation: instead of aggregating over all data points, one can aggregate over each cluster and get a cluster-specific important score (see eq.~\ref{eq:cluster-specific}). For feature $j$ and cluster $k$, let $N_k^{cal} = \sum_{i \in \cI_{cal}} \mathbf{1}(\vz_i = k)$:
\begin{equation}
    \hat \Delta_{j,k}^{\text{split}} = \sum_{i \in \cI_{cal}} \frac{\mathbf{1}(\vz_i = k)}{N_k^{cal}}\bigg(\text{Error}\big(\vz_i, \hat f_{tr}^{-j}(X_{i, -j};(\mX_{tr, -j}, \vz_{tr}) )\big)- \text{Error}\big( \vz_i, \hat f_{tr}(X_{i,:}; (\mX_{tr}, \vz_{tr}))\big)\bigg)
    \label{eq:cluster-specific}
\end{equation}

\noindent Similarly, we can derive the Cluster-specific LOCO-MP estimates for any point-wise error metric, where for feature $j$ and cluster $k$, 
\begin{equation}
    \hat \Delta_{j,k}^{\text{MP}} = \sum\limits_{i=1}^N \frac{\mathbf{1}(\vz_i^* = k)}{\sum_{i=1}^N \mathbf{1}(\vz_i^*=k)}\bigg(\text{Error}\big(\vz_i^*, \hat H^{-j}(X_{i, -j})\big)- \text{Error}\big( \vz_i^*, \hat H(X_{i,:})\big)\bigg)
    \label{eq:cluster-specific-MP}
\end{equation}
This property makes our score highly interpretable at both a \textit{local} and \textit{global} scale which is not the case for most scores in the literature: only the SHAP-derived quantities and LRP admit sample-level to global-level interpretation. For clustering however it is especially important to be able to assess the model at the cluster level, especially when those clusters are used for scientific analysis in downstream tasks. We present such a case in section 4. 

\subsubsection{Choice of hyperparameters}
Cluster LOCO is a family of post-hoc interpretability metrics and is not intended to replace careful selection and validation of the clustering model itself. The choice of clustering algorithm, distance or similarity measure, number of clusters, and other model-specific hyperparameters
should be made before using Cluster LOCO metrics, drawing on appropriate domain knowledge and clustering validation procedures (see \citet{yin_hierarchical_2009,luxburg_clustering_2009,handl_computational_2005,ullmann_validation_2022,gan_are_2025}).
Once a clustering solution has been selected, Cluster LOCO metrics can then be used to interpret which features contribute most to that solution. For Cluster LOCO-Split and Cluster LOCO-MP, the main tuning choices are the train/calibration split ratios and minipatch size ratios respectively. For the former, both splits must contain enough observations to obtain meaningful cluster labels and stable generalizability estimates. We recommend shuffling the data before splitting and using a balanced split, such as a 50\% train and 50\% calibration split, as a default. This choice helps avoid pathological splits in which rare or unbalanced clusters are poorly represented in either the training or calibration set. For Cluster LOCO-MP, the main hyperparameters are the minipatch size ratios. The minipatch size controls the tradeoff between computational efficiency and the accuracy of the score estimates: smaller minipatches are faster but may yield noisier clustering solutions, while larger minipatches better approximate the full-data clustering problem at greater computational cost. The minipatch size ratios can be tuned in general as discussed in ~\citet{gan_model-agnostic_2023} and tend to be robust when selecting the appropriate ratios ($r_N = \frac{n}{N}, r_M = \frac{m}{M}$) in the range of 20\%-50\% for both observations and features \citep{gan_fast_2022}. 

\noindent On the other hand, our scores also require choosing an appropriate transfer classifier. We contend such a classifier should be sufficiently flexible to predict the cluster assignments. In our sensitivity experiments (Figures~\ref{fig:supp_fig5}, \ref{fig:supp_fig6}), classifier choice has relatively little effect on feature rankings in the simpler settings, but becomes more consequential for nonlinearly separable clusters. We therefore recommend a random forest as a practical starting point, particularly when the cluster structure is nonconvex, and checking its baseline generalizability error before interpreting the feature importance scores: a large such error can reflect either an inadequate classifier or poor generalizability of the clustering procedure. For both global and cluster-specific interpretation, we recommend soft classifiers combined with a point-wise error score, such as the multi-class hinge error, so that the same prediction-error differences can be averaged over all observations or within each cluster.

\section{Simulations}
Our Cluster LOCO family is designed to address two limitations of existing clustering feature importance methods: dependence on specific clustering algorithms and limited scalability to large, high-dimensional datasets. In this section, we evaluate our proposed Cluster LOCO-Split and Cluster LOCO-MP/LOCO-RAMPART methods against existing feature importance approaches in simulation settings. We compare our metrics against five baseline methods that provide feature importance for clustering: a prototype-based feature importance (PBFI) and Fuzzy C-Means Shapley (FCM SHAP) from~\citet{napoles_feature_2024}, permutation feature importance (PFI) \citep{pfaffel_featureimpcluster_2020}, an extension of layer-wise relevance propagation (LRP) for KMeans clustering proposed by~\citet{kauffmann_clustering_2024}; and IMPACC, a minipatch-based adaptive consensus clustering algorithm proposed by~\citet{gan_fast_2022}. Since no public code base was available for PBFI or FCM SHAP, we re-implemented both methods in Python. We also extended the KMeans-based PFI implementation, originally proposed in R, to arbitrary \texttt{scikit-learn}-style clustering algorithms, and similarly adapted IMPACC, originally in R as well, to be compatible with these algorithms. Wherever possible, we compare feature importance methods using the same base clustering algorithm. However, the clustering solution itself is not held fixed, as each method requires its own fitting procedure - feature perturbation, data splitting, minipatch sampling etc. Therefore these comparisons assess feature importance under a common base clustering algorithm, rather than for an identical clustering solution.

\subsection{Illustrative simulation for Cluster LOCO-Split}

To illustrate the limitations of existing methods and how Cluster LOCO-Split is able to obtain better feature importance scores in even simple examples, we construct a small but challenging synthetic example for clustering. This synthetic clustering example consists of $3$ signal features created from interlaced half circles with varying levels of overlap and noise shown in Figure~\ref{fig:fig1}a; details on the data-generating process are provided in the Appendix. We augment those $3$ signal features with $2$ pure noise features sampled from independent uniform distributions. The difficulty of this example stems from the nonlinear separability of the clusters, which challenges methods that favor convex cluster geometries.  

\begin{figure}[!ht]
    \centering
    \rowlabel{a}
    \rowcontent{%
        \begin{subfigure}{\linewidth}
            \centering
            \includegraphics[width=\linewidth]{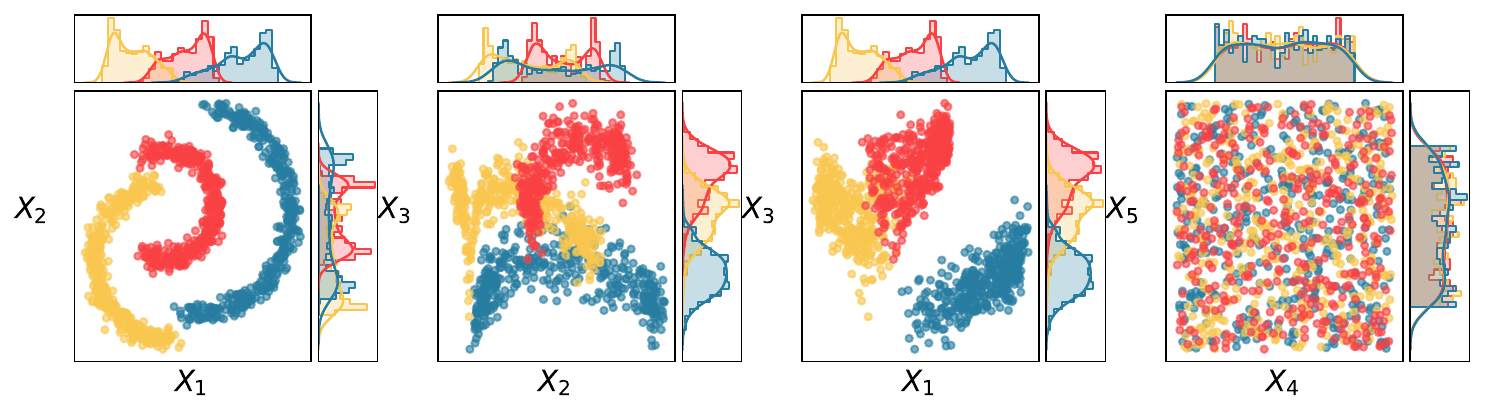}
        \end{subfigure}
    }
    \rowlabel{b}
    \rowcontent{%
        \begin{subfigure}[t]{0.195\linewidth}
        %\scriptsize \textbf{1.}\par\vspace{0.2em}
            \centering
            \includegraphics[width=\linewidth]{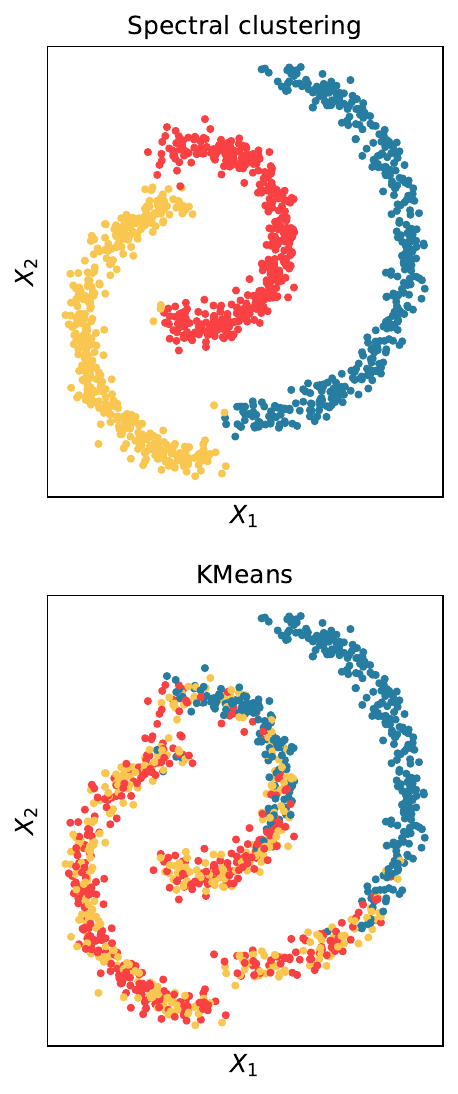}
        \end{subfigure}
        \hfill
        \begin{subfigure}[t]{0.805\linewidth}
        %\scriptsize \textbf{2.}\par\vspace{0.2em}
            \centering
            \includegraphics[width=\linewidth]{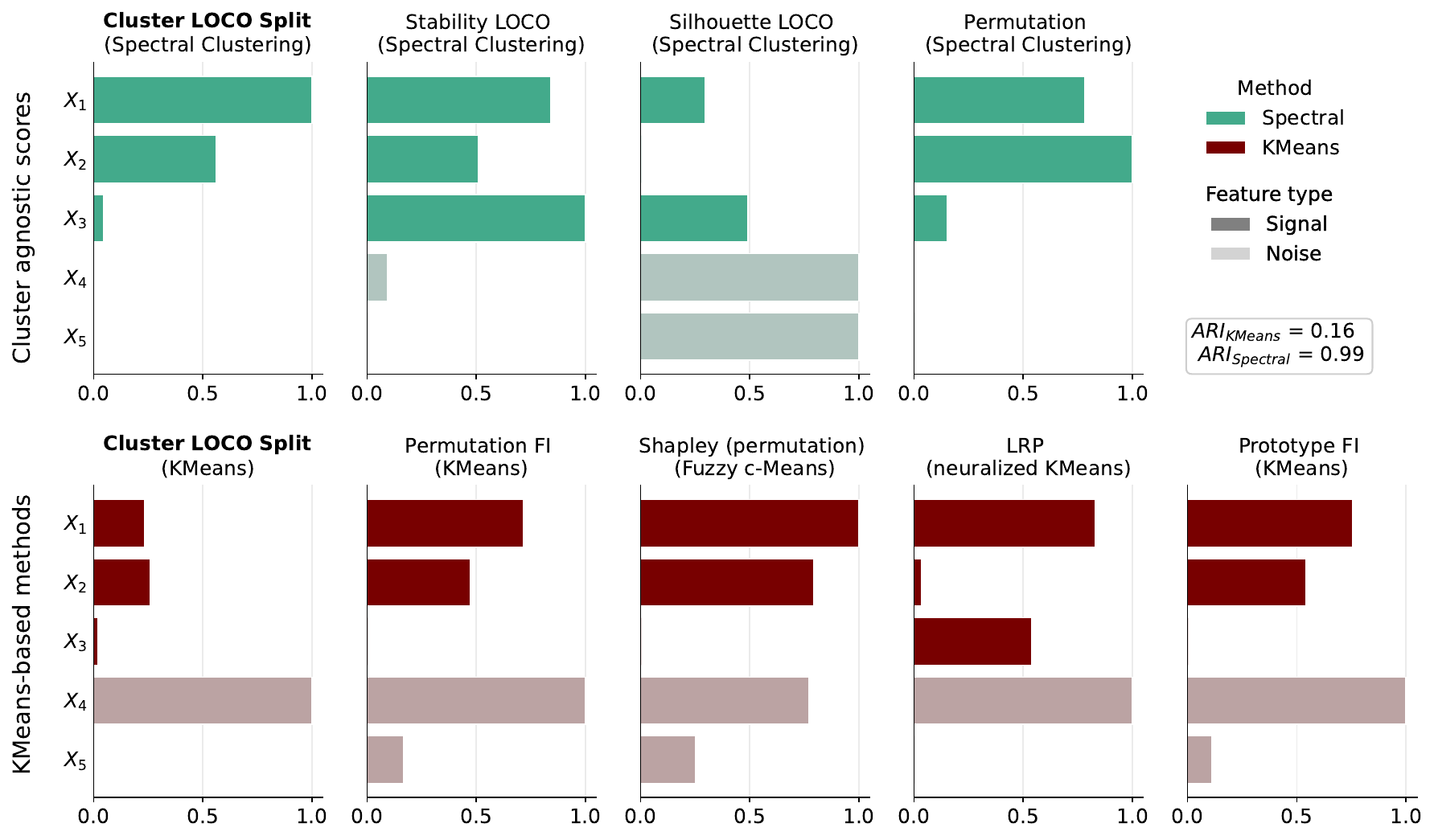}
        \end{subfigure}
    }
    \caption{Cluster LOCO is a reliable feature importance score for complex clustering problems. \textbf{a. Simulated dataset} presents nonlinearly separable clusters, first three features are signal features, ordered by their importance of contribution. Removing $X_1$ leads to a harder clustering problem in the remaining signal space with overlapping nonlinearly separable clusters, removing $X_2$ leaves a moderately hard problem while removing $X_3$ leads to the easier problem hence the ordered importance of each feature. The last two features are pure noise features sampled from a uniform distribution. \textbf{b. Feature importance scores} of proposed and existing IML methods for clustering in the literature. For model-agnostic methods we use Spectral Clustering that yields the highest adjusted Rand index (ARI) on this problem while we report the method used for model-specific algorithms. Our Cluster LOCO importance score is the only metric evaluated able to recover the correct signal features with the correct ordering. \textit{Additionally, we show that interpreting the feature importance of a mis-specified model can be highly misleading with KMeans as baseline.}}
    \label{fig:fig1}
\end{figure}

\noindent  This example shows that choosing the adequate clustering algorithm is crucial before interpreting the features of a clustering solution. Spectral Clustering recovers the true groups as indicated by an adjusted Rand index of nearly 1, whereas methods that impose convex or linearly separable clusters, such as KMeans with an adjusted Rand index of 0.40, produce incorrect assignments as shown in Figure~\ref{fig:fig1}b. Consequently, feature importance methods using KMeans, including FCM-SHAP (where we used exact Shapley), LRP, PBFI, and the original permutation-based score PFI, explain a misspecified clustering model rather than the clustering structure of interest. As shown in Figure~\ref{fig:fig1}b, these methods assign high importance to noise features in this setting. A model agnostic feature importance method for clustering is here crucial to correctly infer feature importance. Therefore we investigated several LOCO-style scores based on different clustering validation criteria: we show two such variants constructed from popular scores, silhouette LOCO is based on the silhouette score change when removing one feature, stability LOCO computes the change in negative ARI under the model-explorer stability objective, using the framework of \cite{ben-hur_stability_2001}. We find that our generalizability-based score, Cluster LOCO-Split, provides the most informative notion of feature importance in this simple yet challenging setting where it uniquely separates signal from noise and correctly orders the signal features by importance. 

\begin{figure}[htbp]
    \centering

    \begin{subfigure}[t]{0.325\textwidth}
        \vspace{0pt}
        \centering
        \begin{tcolorbox}[simbox]
            \centering
            \vspace{1em}
            \panelimg{A1.}{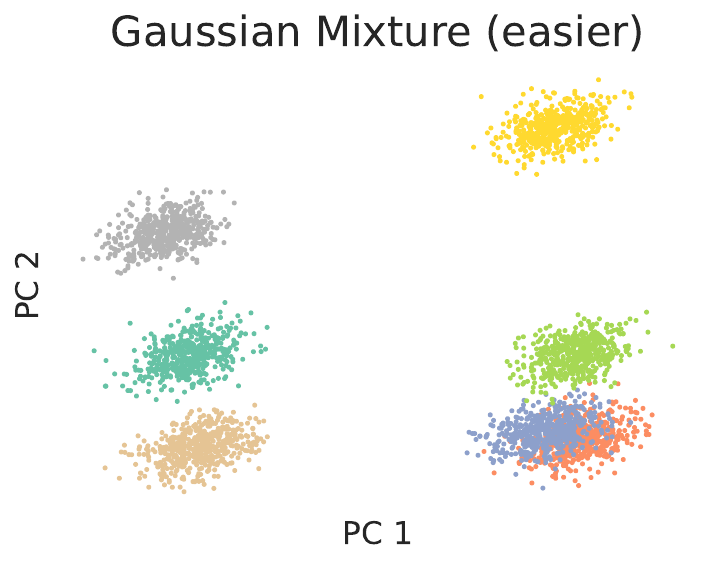}

            \vspace{1em}
            \panelimg{A2.}{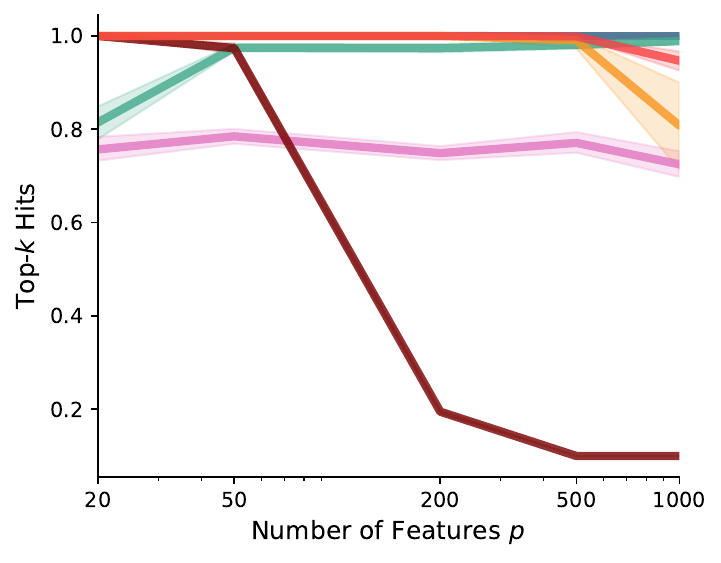}

            \vspace{1em}
            \panelimg{A3.}{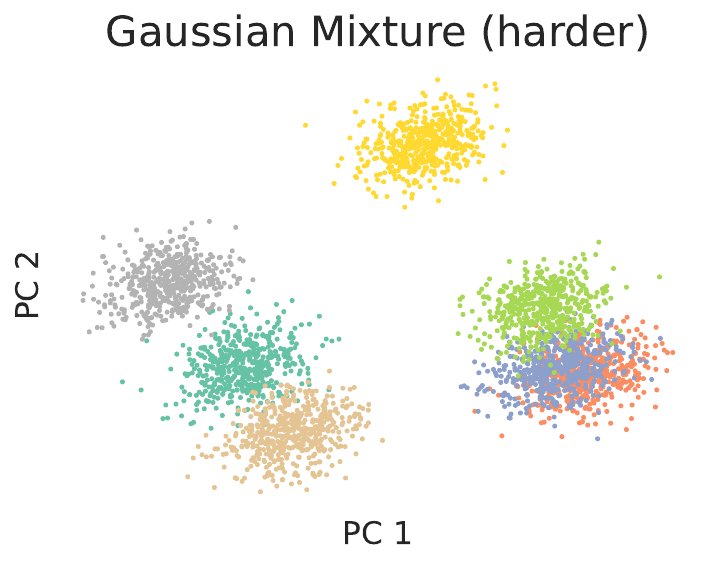}

            \vspace{1em}
            \panelimg{A4.}{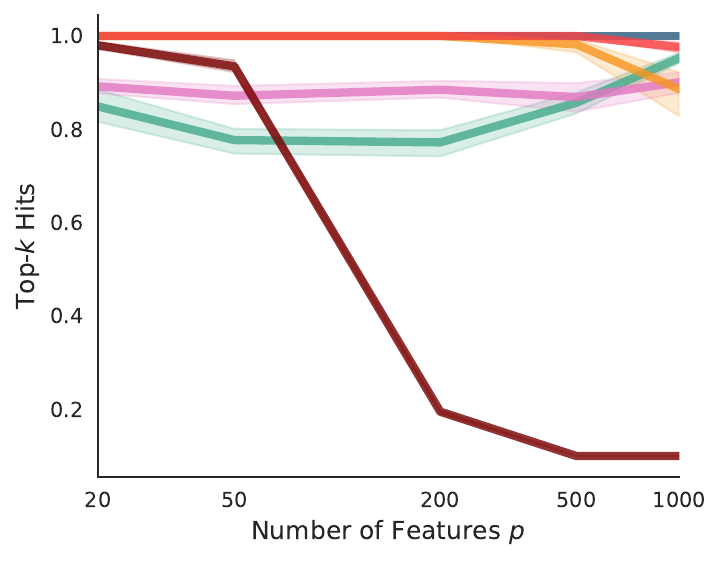}
        \end{tcolorbox}
    \end{subfigure}
    \hfill
    \begin{subfigure}[t]{0.325\textwidth}
        \vspace{0pt}
        \centering
        \begin{tcolorbox}[simbox]
            \centering
            \vspace{1em}
            \panelimg{B1.}{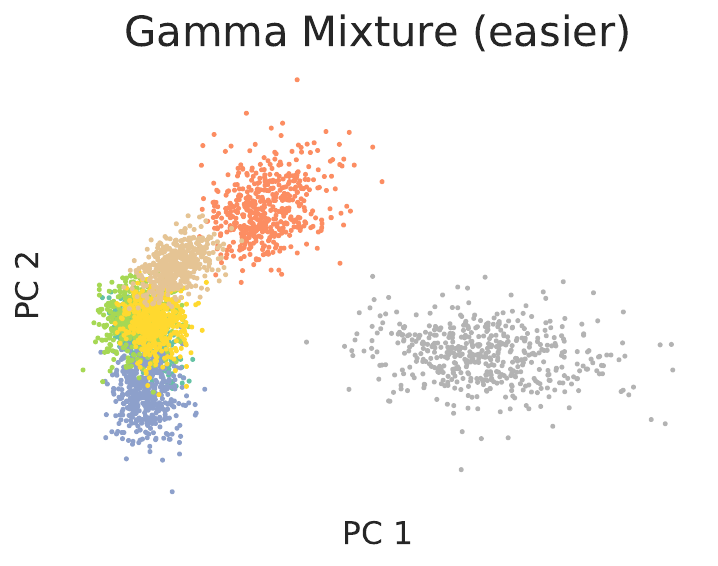}

            \vspace{1em}
            \panelimg{B2.}{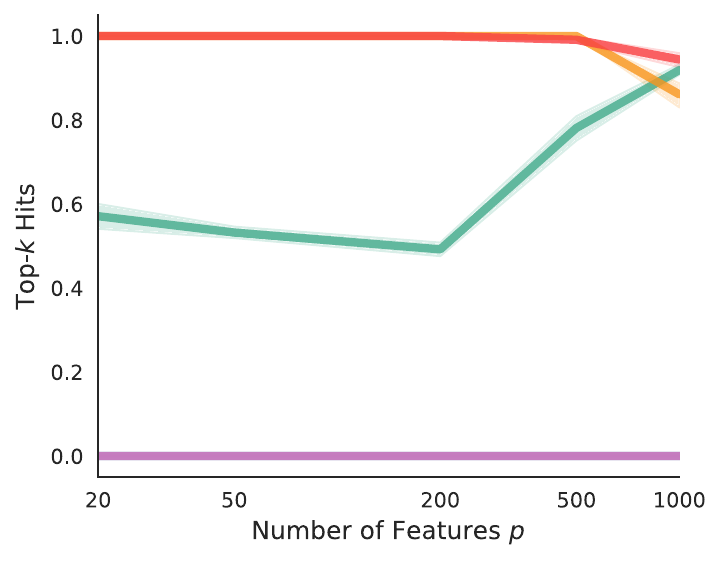}

            \vspace{1em}
            \panelimg{B3.}{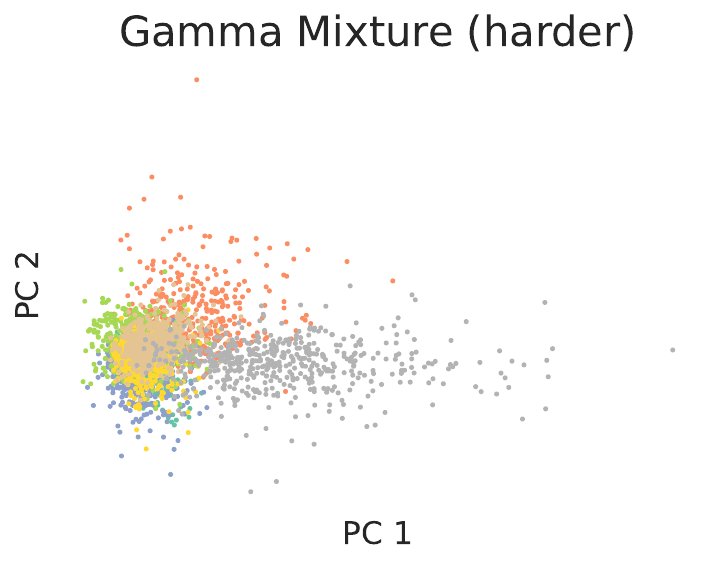}

            \vspace{1em}
            \panelimg{B4.}{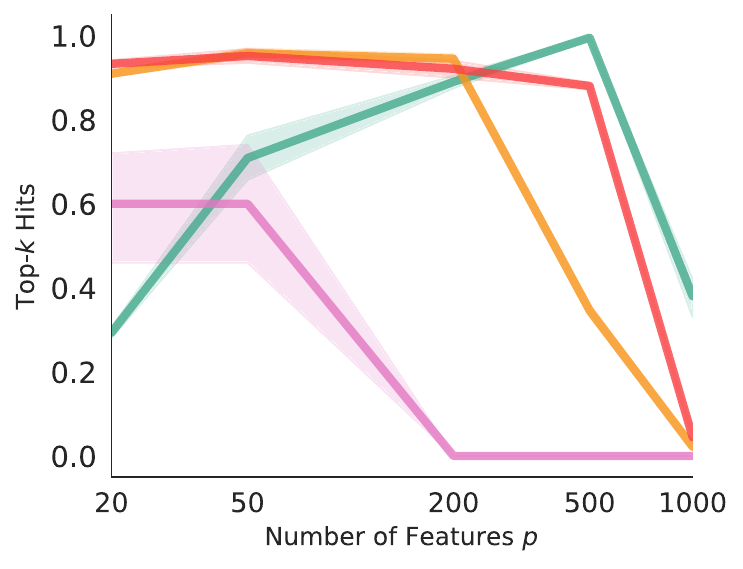}
        \end{tcolorbox}
    \end{subfigure}
    \hfill
    \begin{subfigure}[t]{0.325\textwidth}
        \vspace{0pt}
        \centering
        \begin{tcolorbox}[simbox]
            \centering
            \vspace{1em}
            \panelimg{C1.}{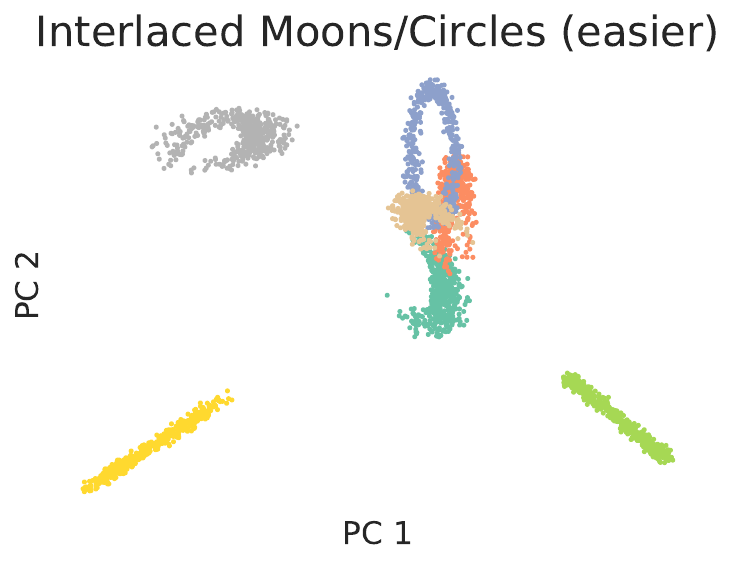}

            \vspace{1.3em}
            \panelimg{C2.}{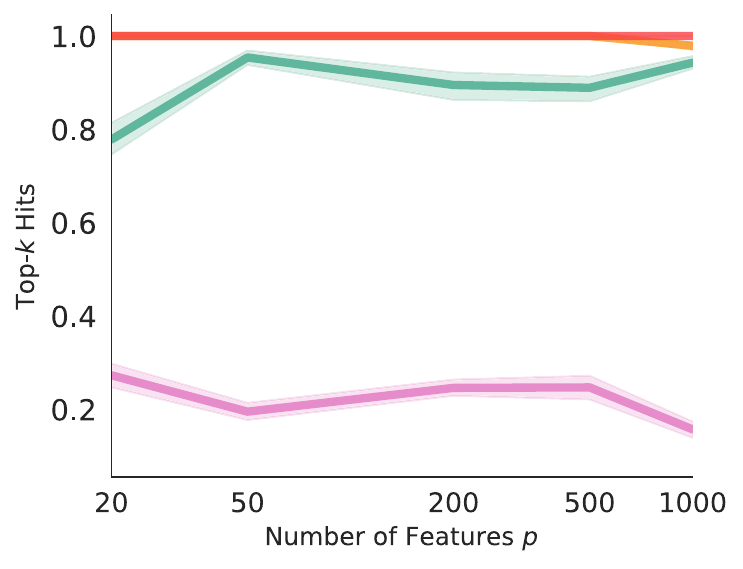}

            \vspace{1.3em}
            \panelimg{C3.}{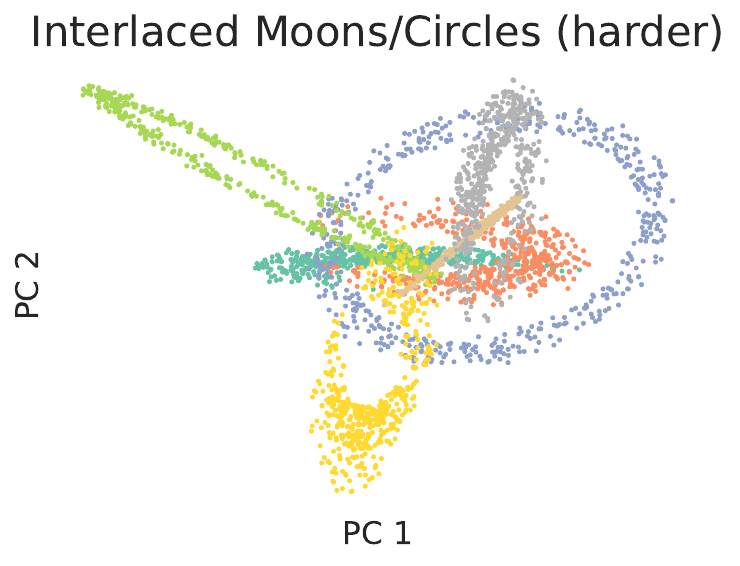}

            \vspace{1.25em}
            \panelimg{C4.}{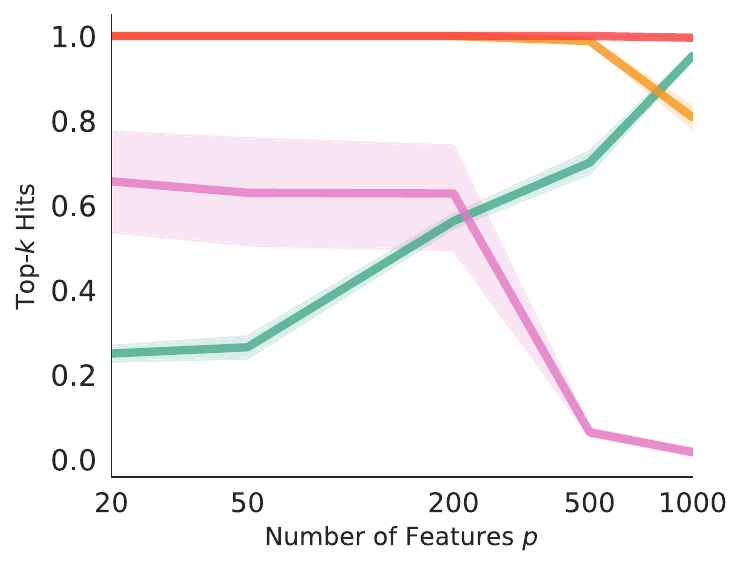}
        \end{tcolorbox}
    \end{subfigure}

    \begin{subfigure}[t]{\textwidth}
        \centering
        \includegraphics[width=\linewidth]{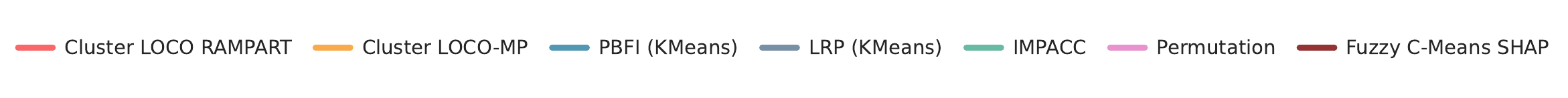}
    \end{subfigure} 
    \caption{\textbf{Simulation results} are obtained across three settings: \textbf{A. Gaussian mixture}, \textbf{B. Gamma mixture}, and \textbf{C. interlaced moons/circles}. Model-agnostic algorithms used are KMeans for Gaussian mixtures, Gamma mixture EM for Gamma mixture, Spectral Clustering for interlaced moons/circles.  For each setting, \textbf{rows 1 and 3} show the data colored by true labels in PC space, while \textbf{rows 2 and 4 }report top-10 hits for feature importance with signal features $p^* = 10$ and noise features $p_{\text{noise}} \in \{10, 40, 190, 490, 990\}$. Model-specific feature importance scores' models are reported in the legend. Top-$k$ hits were averaged over 100 replicates. Cluster LOCO-MP and RAMPART outperform existing methods on most tasks.}
    \label{fig:fig2}
\end{figure}

\subsection{Cluster LOCO for large data via minipatch ensembles}

We next applied Cluster LOCO-MP for large data: generating three base clustering simulations with $K=7$ clusters, $N = 3500$ observations ($500$ observations per cluster), with fixed signal features $p^* = 10$ and increasing noise features $p_{\text{noise}} \in \{10, 40, 190, 490, 990\}$ covering different clustering difficulties. We propose two mixtures models: a Gaussian mixture with \textit{onion} covariance structure (as proposed by \citet{qiu_generation_2006}) to introduce correlation in the features and a Gamma mixture model that generates data with heavier tails. For a more complex scenario of clustering, we generate clusters inspired by the two-dimensional toy examples of moons and concentric circles \citep{ester_density-based_1996}: we sample our data in a two-dimensional embedding with controlled geometry and project up using an orthogonal projection to get any higher-dimensional extension. Each of those three base simulations has two difficulty levels created by varying cluster separation or signal-to-noise ratio into easier or harder clustering problems as shown in the PC-space scatter plot of the generated data in Figure~\ref{fig:fig2} row 1 and 3. We report the top-$10$ hits (or precision at $10$ features) of the true $10$ signal features for each method i.e. $\frac{1}{10}|\{\text{top } 10 \text{ features by method}\} \cap \{\text{true 10-feature set}\}|$. Our Cluster LOCO-MP and Cluster LOCO RAMPART methods were run with fixed $B=5000$ and $B_{\text{rampart}}=1000$ and minipatch ratio sizes $\alpha_N = 0.2, \ \alpha_M = 0.2$, while IMPACC was run with the default hyperparameters. Because FCM SHAP is extremely computationally expensive to run for exact Shapley value, we used in this example the SHAP permutation approximation of Shapley values with 20 iterations. For each simulation setting, model-agnostic methods were evaluated with the best performing algorithm among typical clustering algorithm of the \texttt{scikit-learn} environment: for Gaussian mixtures, we used KMeans, for the interlaced moons/circles, we used Spectral Clustering. For Gamma mixture, we implemented an EM Gamma mixture model that was best suited for clustering this type of data.

\noindent  In Figure~\ref{fig:fig2} panels A2 and A4, the performance of the different feature importance scores in the Gaussian mixture model case is reported via the top-10 hits: PBFI and LRP which are KMeans-based recover the correct features as expected, our Cluster LOCO-RAMPART and Cluster LOCO-MP scores also recover the correct top-10 features for $p \leq 500$, with a slight drop in performance at $p=1000$ which could be solved by budgeting ahead more minipatches. In the easy case, IMPACC performs well, but improves especially when the number of noise features grows as it is designed for sparse features. FCM SHAP and PFI perform the worst overall in the easy and hard case of the Gaussian mixture. For FCM SHAP, the steep drop off of performance is due to the gap between exact Shapley and the approximation with fixed number of iterations while the number of features grow. Hence, despite not being KMeans-specific, Cluster LOCOMP and RAMPART achieve similar performance with the allocated budget in terms of feature importance recovery.

\noindent In the two other simulation examples in Figure~\ref{fig:fig2} columns B and C, Cluster LOCO-MP and Cluster LOCO-RAMPART achieve the highest or comparable top-10 recovery across most of these settings: in the Gamma mixture case, IMPACC achieves similar or better performance only when the features are extremely sparse (5\% or less of the features are signal features), while in the easy case, both Cluster LOCO RAMPART and Cluster LOCO-MP have robust recovery of the top-10 features. We see similar results with the interlaced moons/circles (Figure~\ref{fig:fig2} panels C2 and C4), Cluster LOCO-MP and Cluster LOCO RAMPART being extremely consistent and effective at recovering the true signal features. Overall, Cluster LOCO methods for large data perform as well as or outperform existing methods for those diverse clustering simulations, even typically "hard" clustering tasks, and we provide further simulation results in the Additional figures.

\section{An application in single-cell transcriptomics to immune cells}

We apply Cluster LOCO-MP and Cluster LOCO-RAMPART for scientific discovery in a real-world clustering application. In single-cell transcriptomics, clustering is routinely used to discover putative cell types or states, after which differentially expressed genes are identified post-hoc and interpreted as marker genes for the discovered groups. This workflow is embedded in popular pipelines such as Scanpy and Seurat \citep{wolf_scanpy_2018, hao_integrated_2021}, and sometimes considered the gold standard for discovering cell-type identities \citep{luecken_current_2019}. However, this two-step procedure raises important statistical concerns: the same data are first used to define clusters and then reused to test for genes that distinguish those clusters, a form of \textit{double dipping} or \textit{data snooping} known to inflate false discovery rates \citep{lahnemann_eleven_2020}. Moreover, this sort of cell-type clustering for gene marker annotation is not a unified framework: the clustering algorithm choice can substantially change the downstream discoveries, and those new marker genes are implicitly dependent on the clustering solution, which can contribute to a lack of reproducibility of results \citep{gibson_perspectives_2022}. Cluster LOCO addresses this gap by directly quantifying the contribution of each feature to clustering generalizability, providing a clear unified framework for feature importance in clustering rather than relying on separate downstream testing steps.

\subsection{Peripheral Blood Mononuclear Cells dataset for immune cell types discovery} 

We study the Peripheral Blood Mononuclear Cells (PBMC) dataset published by~\citet{zheng_massively_2017} that was annotated using purified transcriptomics populations. Our goal is to cluster and understand the cluster-wise marker genes discovered using traditional pipelines of gene annotations (known to suffer from double dipping) and our proposal of feature importance. We use the processed data available via Scanpy \citep{wolf_scanpy_2018} that consists of 765 genes and 700 single cells, obtained after normalizing and scaling the data as reported on the 10X Genomics repository. The reported labels were obtained by classifying the single-cell data with 11 purified sub-populations of PBMC reference profiles and are referred to as \textit{bulk labels} or \textit{purified labels}. It is important here that we chose data that was not generated via clustering but via a correlation approach for external validation of our results. To further illustrate our results, we compiled known marker annotations of human PBMCs from three sources: the Azimuth atlas \citep{stuart_comprehensive_2019}, as well as~\citet{ding_systematic_2020} and~\citet{oelen_single-cell_2022} supplementary data on known markers of human PBMC cell types. When pooling known markers, we note that the granularity of cell typing was not always consistent, and this leads to some markers ambiguously marking multiple cell types.

\subsection{Globally important genes for clustering}

\begin{figure}
    \begin{subfigure}[t]{0.32\linewidth}    \noindent\textbf{a.}\par\vspace{0.2em}
            \centering
     \includegraphics[width=\linewidth]{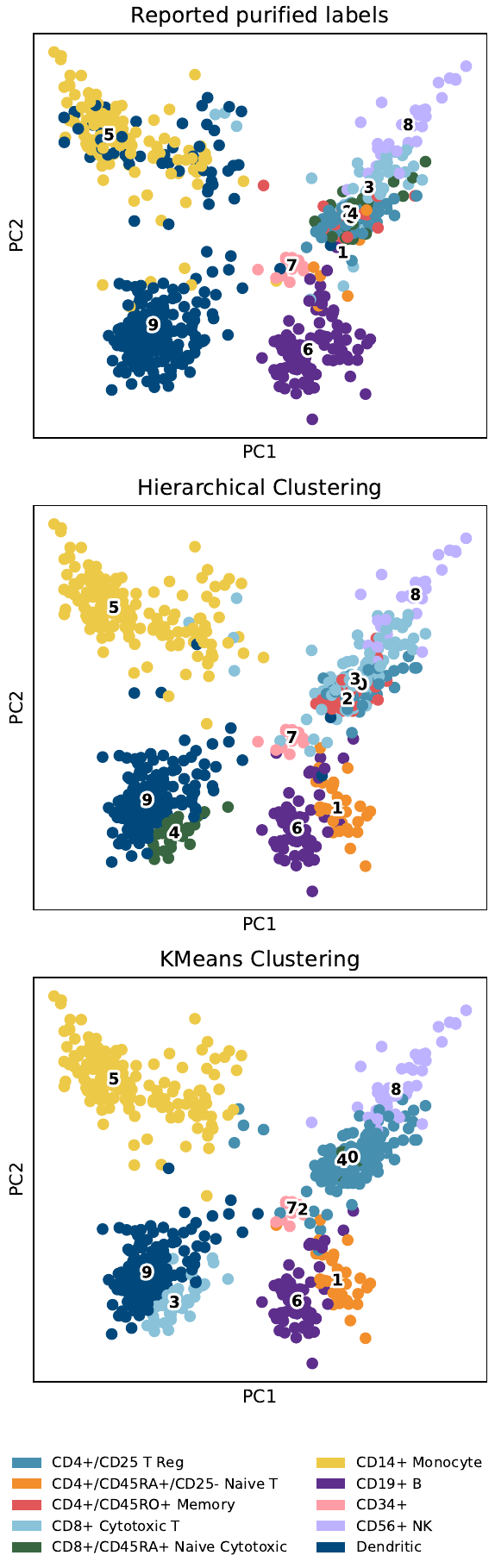}
        \end{subfigure}
        \hfill
        \begin{subfigure}[t]{0.679\linewidth}
    \noindent\textbf{b.}\par\vspace{0.2em}
            \centering
     \includegraphics[width=\linewidth]{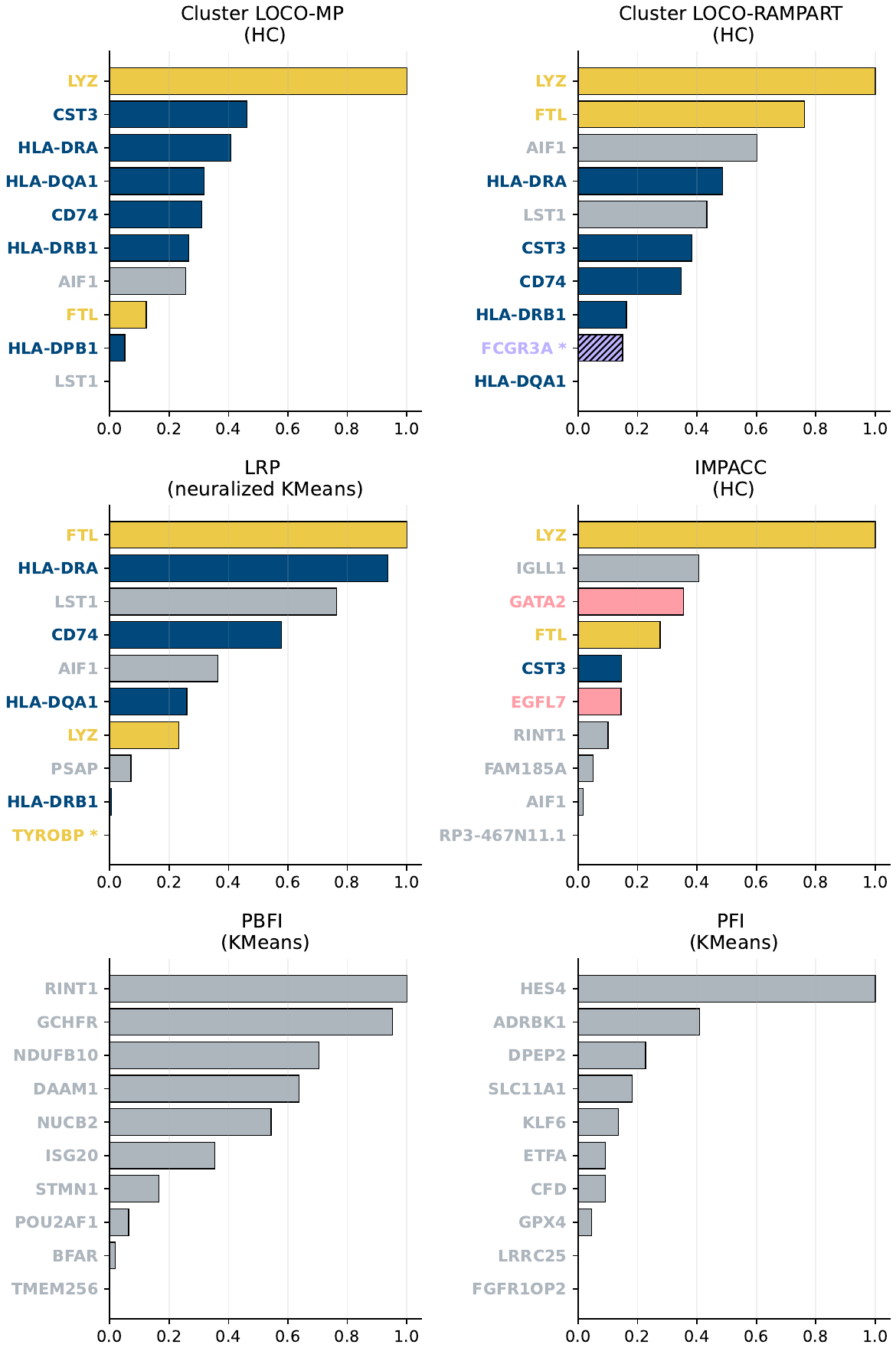}
        \end{subfigure}
\caption{\textbf{Top 10 most globally important genes} obtained via different feature importance methods. \textbf{a. Normalized gene expression}, reported and clustered via base clustering models - Hierarchical Clustering and standard KMeans clustering, in PC space. \textbf{b. Feature importance scores}: we color the genes by known marker genes extracted from the Azimuth Human PBMC database \citet{hao_integrated_2021} and the manual annotations from \citet{ding_systematic_2020} and \citet{stuart_comprehensive_2019}. Genes that appear not to be in any known PBMC cell type marker set are colored in grey. We note that Cluster LOCO-MP finds the most marker genes, KMeans-based methods fail to identify known markers in the important genes.}
\label{fig:fig3}
\end{figure}

In this data application, global feature importance scores identify the genes most influential for the overall clustering solution. Because the reported labels provide external biological annotations, we use known marker genes to assess whether feature importance recovers biologically annotated features when the underlying clustering solution approximately recovers these cell populations. 
We report in Figure~\ref{fig:fig3} the results for six clustering feature importance metrics: Cluster LOCO-MP, Cluster LOCO-RAMPART with top-100 genes,  IMPACC (consensus clustering with minipatches importance), LRP (layer-wise relevant propagation score), PBFI (prototype-based feature importance) and PFI (permutation feature importance). We chose hierarchical clustering with Ward linkage with $K=10$ clusters to correspond with the 10 known purified labels as base clustering algorithm for our model-agnostic feature importance scores. We show in Figure~\ref{fig:fig3}a the clustering solutions in PC-space aligned with the reported labels for both hierarchical clustering and KMeans clustering that PBFI and PFI use. Cluster LOCO-MP was run with $B=5000$ minipatches and minipatch ratio sizes $\alpha_N =0.42$ for observations, $\alpha_M = 0.26$ for features. We aligned the obtained labels for each methods with the reported \textit{purified} labels using the Hungarian algorithm.

\noindent To evaluate the different clustering feature importance methods, we first analyze the top 10 most globally important features in Figure~\ref{fig:fig3}b, where the highest-ranked features correspond to the most important genes driving the clustering structure. We first note that KMeans-based importance methods PBFI and PFI fail to find any known marker genes among the 10 most important genes. Among the remaining four methods, the CD14+ monocyte marker \textit{FTL} is consistently important for Cluster LOCO-MP, Cluster LOCO-RAMPART, LRP and IMPACC, matching the original empirical findings in~\citet{zheng_massively_2017}. Cluster LOCO-MP showed greater overlap with the compiled marker reference in this dataset under the evaluated settings, assigning high importance scores primarily to known marker genes from dendritic cells and monocytes, which correspond to the two most well-separated clusters in the PC space. On the other hand, LRP that uses a neuralized KMeans and therefore is more flexible than traditional KMeans recovers similar reference markers as Cluster LOCO-MP. In particular, Cluster LOCO-MP and LRP share important genes like \textit{AIF1} and \textit{LST1} that do not belong in our reference marker set but whose expression patterns have been documented in the monocyte and myeloid cell types in the PBMC literature \citep{thul_human_2018, leon-oliva_aif1_2023, ferreira_identifying_2024}. However, LRP also presents \textit{PSAP} a gene with lower immune-lineage specificity \citep{thul_human_2018}, and therefore with a less interpretable profile for cell typing. We note that IMPACC’s top-ranked features include a mixture of CD34+ hematopoietic stem cell markers and monocyte markers. However, half of its top 10 most important genes lack documented cell-type specificity (e.g., \textit{IGLL1}). These results underscore that in order to interpret important genes, the clustering solution needs to match a relevant biological truth. In this example, the hierarchical clustering solution agrees more closely with the reference annotations, providing a more suitable basis for cell-type interpretation than KMeans clustering assignments.

\subsection{Cluster-wise important genes and marker selection} 
\begin{figure}[htbp]
    \centering
    \labeledrowcontent{a.}{\linewidth}{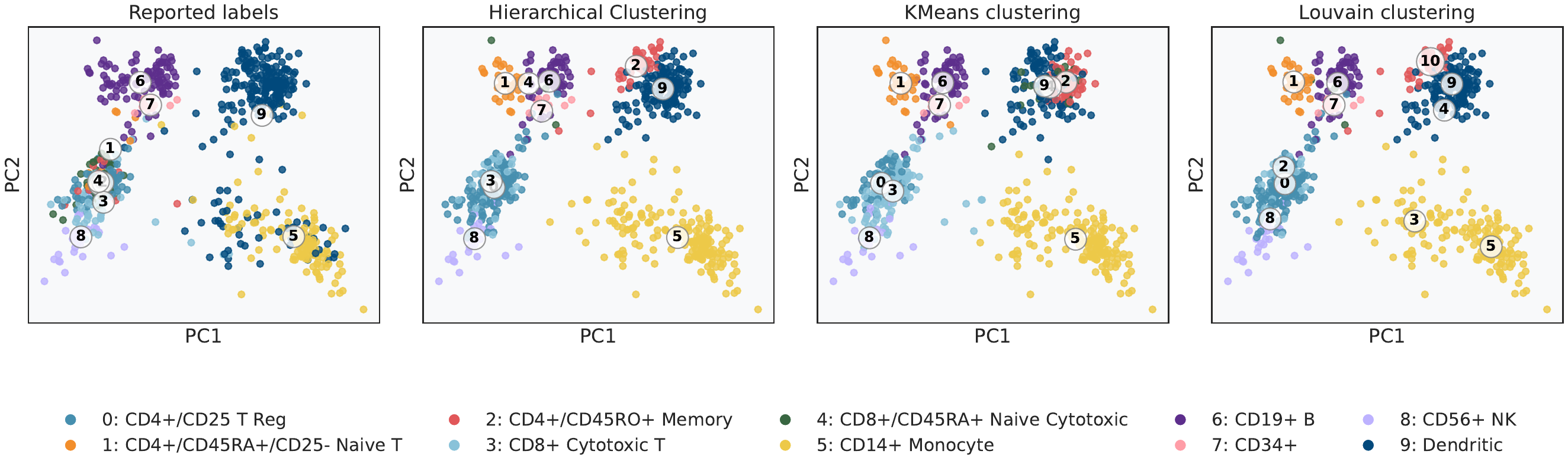}{0.03\linewidth}
    \vspace{0.2em}
    \labeledrowcontent{b.}{1.1\linewidth}{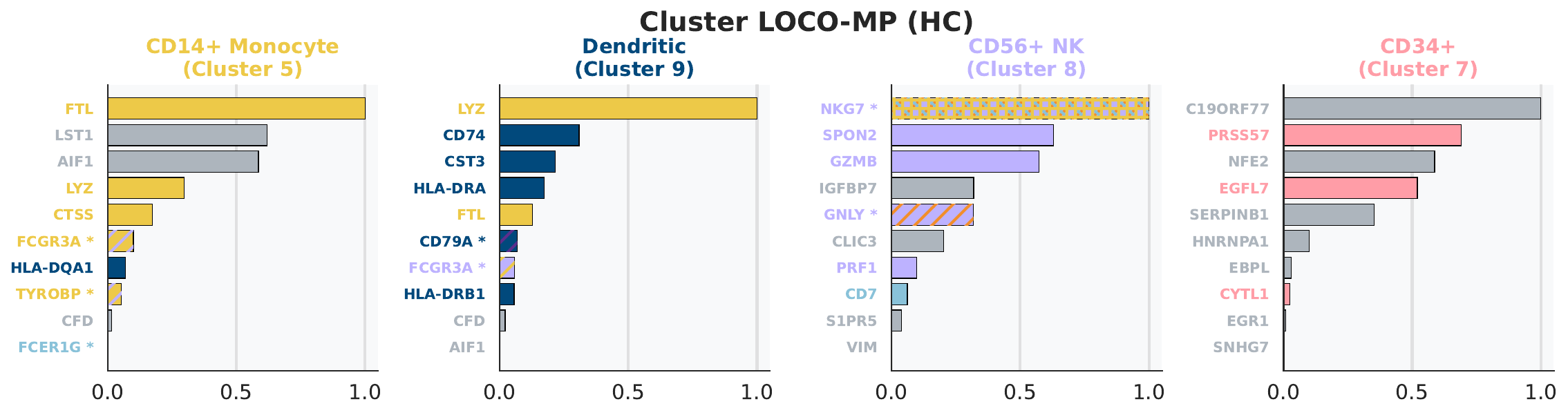}{-0.05\linewidth}
    \vspace{0.2em}
    \labeledrowcontent{c.}{1.1\linewidth}{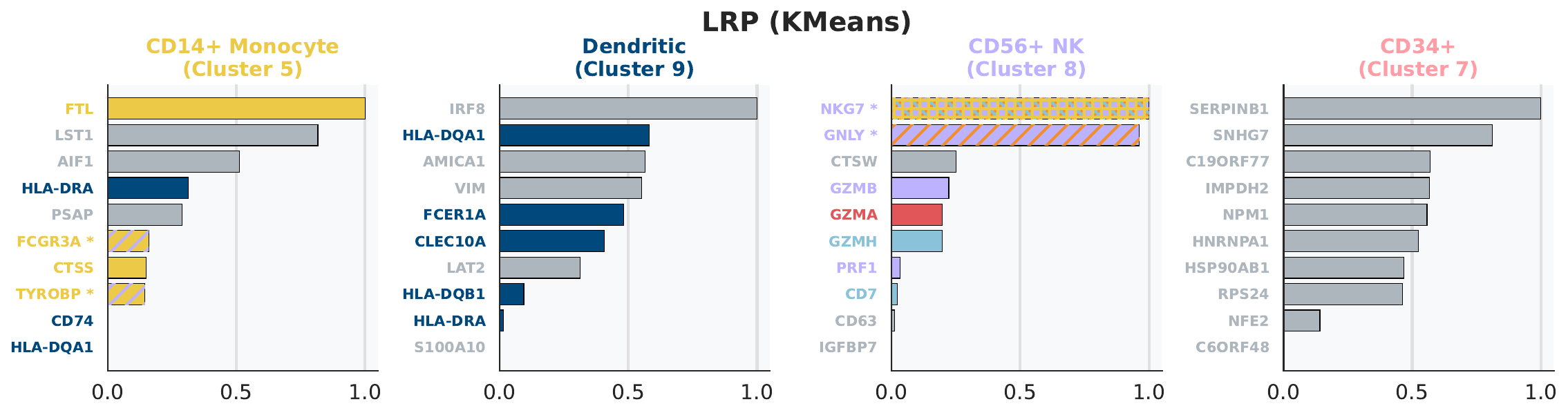}{-0.05\linewidth}
    \vspace{0.2em}
    \labeledrowcontent{d.}{1.1\linewidth}{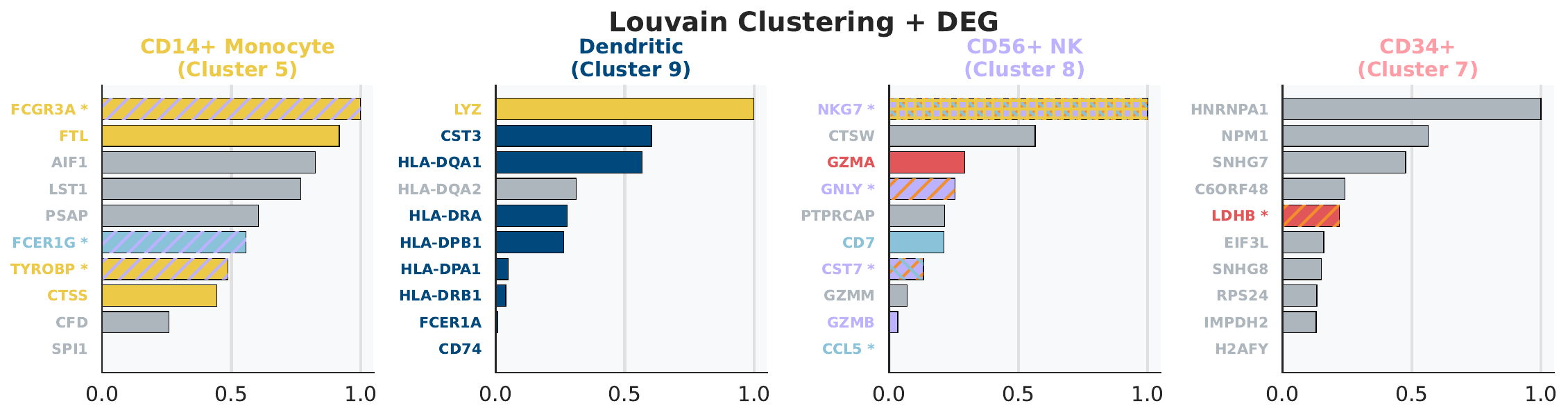}{-0.05\linewidth}
    \caption{\textbf{Top 10 most important genes for clusters} corresponding to well separated cell types: CD14+ Monocytes (Cluster 5), Dendritic cells (Cluster 9), Natural Killer cells (Cluster 8) and CD34+ hematopoietic stem cells (Cluster 7).  \textbf{a. Normalized gene expression} in PC space colored by cluster labels aligned with the reported labels. KMeans labels are from the neuralized KMeans model. \textbf{b. Cluster LOCO-MP} feature importance scores: Hierarchical Clustering was used as base clustering algorithm. 
    \textbf{c. Layer-wise Relevant Propagation (LRP)} feature importance scores for neuralized KMeans.   \textbf{d. Louvain clustering and differential gene expression}: genes ranked in size effect among those tested significant at level 0.05 in differential expression testing. Our method Cluster LOCO-MP captures more consistent cluster-specific markers.}
    \label{fig:fig4}
\end{figure}

Global feature importance scores identify the genes most influential for the overall clustering solution, but cluster-level scores are needed to assess which genes characterize individual cell types. We compare here Cluster LOCO-MP at cluster-level with LRP also aggregated at cluster-level -- other importance scores investigated earlier are solely global feature importance scores. In parallel, we implement the \textit{clustering + differential gene expression} workflow commonly used in single-cell analysis and described in Scanpy tutorials \citep{wolf_scanpy_2018}: data is clustered using Louvain clustering, identifying 11 clusters, and subsequent differential gene expression is obtained using t-test across Louvain clusters with Benjamini-Hochberg correction. We define differentially expressed genes (DEGs) using a 5\% adjusted p-value cutoff and rank significant genes by the magnitude of their z-scores. We show in Figure~\ref{fig:fig4}a the data in PC-space colored with the respective methods, notably hierarchical clustering for Cluster LOCO-MP, neuralized KMeans for LRP and Louvain clustering for differential gene expression where labels have been aligned with the reported \textit{purified} labels. Cluster LOCO-MP was also run with $B=5000$ minipatches and minipatch ratio sizes $\alpha_N =0.42$ for observations, $\alpha_M = 0.26$ for features for the cluster-level analysis. \\ 

\noindent The cluster-level feature importance are here analyzed for three well separated cell types (CD14+ monocytes, dendritic cells, CD56+ natural killer cells) and one cell type that is harder to cluster (CD34+ hematopoietic stem cells) shown in the PC space in Figure~\ref{fig:fig4}a. For each aligned cluster, we compare the top-ranked genes from Cluster LOCO-MP, LRP, and the standard Scanpy workflow (Figure~\ref{fig:fig4}b--d), with full top-10 rankings for all cell-type-aligned clusters reported in the Additional figures (Figures~\ref{fig:supp3a},~\ref{fig:supp3b},~\ref{fig:supp3c}). Across all clusters, Cluster LOCO-MP recovers a larger number of cluster-specific marker genes than the competing methods. In particular, for the well separated clusters, Cluster LOCO-MP and LRP produce broadly similar rankings, and both are largely consistent with the DEG-based rankings. However, LRP and DEGs rank several genes with weaker cell-type specificity or genes absent from our marker reference set (\textit{CFD, VIM, LAT2, SPI1, S100A10, PTPRCAP}). We also note that Cluster LOCO-MP and LRP scores for monocytes and dendritic reflect the mixing of true dendritic cells with monocytes as shown in the reported labels of Figure~\ref{fig:fig3}a, with the presence of shared markers. On the other hand, this cell type is largely over-clustered via Louvain (see Figure~\ref{fig:supp3c}) where the DEGs of cluster 4, 9 and the unmatched cluster 10 are very similar. In this particular case study, DEGs show possible false positive signal, with larger amount of markers significant for clusters that do not match the cell type they identify (see table~\ref{tab:tab2}). For the CD34+ hematopoietic stem cell cluster, which is less cleanly separated than the monocyte, dendritic-cell, and NK clusters, Cluster LOCO-MP is largely more interpretable than LRP and differential gene expression analysis. In this case, Cluster LOCO-MP ranks three known CD34+ HSC markers among its top 10 genes, namely \textit{PRSS57}, \textit{EGFL7}, and \textit{CYTL1}, whereas LRP and differential gene expression rank no known CD34+ HSC markers, in fact \textit{LDHB} appears as a DEG although reported to rather be a marker for Naive T cells or Memory T cells. This pattern is consistent with the geometry of the learned clustering shown in the PC space: well-separated clusters yield more specific and interpretable markers, while less cleanly separated populations, particularly T-cell subtypes, lead to more ambiguous marker rankings across all methods. However this validation against known biological markers is necessarily incomplete: genes absent from a reference marker set may reflect noise or poor specificity, but they may also represent missing markers, or previously under-characterized context-specific genes. Additionally, alignment of important genes to biological markers under a suitable clustering algorithm is an assumption, reasonable in the context of the common practices in single cell but not necessarily biologically realistic or true. Nevertheless, Cluster LOCO-MP most consistently prioritizes cluster-specific markers, suggesting that its local feature importance scores better reflect the features supporting the learned clustering structure. More broadly, our analysis with Cluster LOCO-MP provides an alternative perspective on the use of clustering for cell-type marker discovery. In standard single-cell workflows, cluster-derived annotations are commonly treated as fixed labels in subsequent analyses, despite having been learned from the same data. This data-derived nature of the annotations is important for the validity and interpretation of downstream conclusions. In our setting, this dependence is made explicit: the identified important genes characterize the learned clustering solution and their interpretation therefore remains tied to the adequacy of the clustering model used to define the cell populations.

\section{Discussion}

Clustering is often subject to debate regarding its rigor: \textit{is clustering an art or science} \citep{luxburg_clustering_2012}? In this work, we contribute to making clustering more scientifically rigorous by addressing a gap in the interpretability of clustering solutions. We devised Cluster LOCO, a novel family of feature importance metrics for clustering that are model-agnostic, flexible and scalable. Through synthetic and real-world evaluations, we have shown our method's flexibility and Cluster LOCO provides feature-level explanation improving the reliability and trustworthiness of clustering solutions, especially for downstream analysis. In particular, in our single-cell transcriptomics application, we addressed a current challenge in single-cell analysis by providing a new unified framework for gene marker annotation that informs practitioners with interpretable, cluster-specific genes corresponding to cell-types when the clustering algorithm identifies a cluster aligned to the biological cell states.

\noindent Furthermore, Cluster LOCO's model-agnostic design extends beyond typical clustering algorithms to encompass entire clustering pipelines: our framework can be applied on workflows that incorporate preliminary dimensionality reduction, or truly blackbox models, provided they can be applied to obtain new clustering solutions. Cluster LOCO was therefore implemented as an open-source Python library compatible with \texttt{scikit-learn}-style estimators and requires standard \texttt{fit} and \texttt{predict} functionality from the clustering model. As such, Cluster LOCO is universally applicable to both simple algorithms and sophisticated unsupervised architectures, while retaining a straightforward feature interpretation. However, several limitations remain: first, although the minipatch formulation improves scalability, Cluster LOCO-Split still requires fitting a full transfer classifier and one reduced classifier per feature, while Cluster LOCO-MP requires clustering and classifier fitting across many minipatches. These computations can make the methods more demanding than simpler post-hoc scores. For instance, in settings involving neural network-based clustering models, architecture-specific interpretability methods may be more computationally efficient, although typically less general than our proposed approach. Second, Cluster LOCO focuses on feature-level interpretability: it explains which variables are important for a clustering solution globally or cluster-wise, but it does not directly provide sample-level local explanations for why a particular observation was assigned to a particular cluster. 
%%%  Future directions, open questions

\noindent Future work could extend Cluster LOCO along both methodological and computational directions. Computationally, future work could further improve the efficiency of the minipatch procedure through adaptive sampling designs (e.g. multi-armed bandits, or active learning methods). Methodologically, the notion of generalizability could be adapted to unsupervised tasks beyond clustering, including dimensionality reduction, topic modeling, or representation learning. In each case, LOCO-style feature importance would ask which features are necessary for the learned structure to remain stable and generalizable under refitting, providing a path toward feature-level interpretability for a broader class of unsupervised workflows. Finally, an important direction is to further investigate the use of Cluster LOCO in single-cell transcriptomics. Importantly, our single-cell analysis approaches marker identification from an interpretable machine learning perspective and therefore targets a different question from conventional \textit{cluster-then-test} pipelines. Rather than treating cluster assignments as fixed annotations for subsequent inference, Cluster LOCO explicitly interprets them as outputs of a chosen clustering model and quantifies which genes support the generalizability of those assignments. This distinction also emphasizes that clustering should not be viewed solely as a preprocessing or annotation step: when cluster-derived labels inform downstream scientific conclusions, the choice and adequacy of the clustering model remain part of the analysis. Accordingly, Cluster LOCO scores should be interpreted as feature importance for a learned clustering solution, rather than as evidence of causal effects or as a replacement for statistical inference on differential expression. Moreover, our immune-cell analysis provides one case study in which cell-type labels were available from external validation experiments, but future work could evaluate the framework across additional datasets, cell-type annotation settings, pre-processing pipelines, and biological contexts. In particular, it would be useful to study how Cluster LOCO compares with marker discovery pipelines based on reference-atlas cell-type classification. More broadly, we believe Cluster LOCO can support a more transparent and statistically grounded use of clustering in scientific discovery.

\subsubsection*{Code availability and data accession}
Code for Cluster LOCO and scripts for reproducing the main results accompanied with data accession instructions are available at \href{https://github.com/DataSlingers/ClusterLOCO}{https://github.com/DataSlingers/ClusterLOCO}.

% \subsubsection*{Broader Impact Statement}
% In this optional section, TMLR encourages authors to discuss possible repercussions of their work,
% notably any potential negative impact that a user of this research should be aware of. 
% Authors should consult the TMLR Ethics Guidelines available on the TMLR website
% for guidance on how to approach this subject.

% \subsubsection*{Author Contributions}
% If you'd like to, you may include a section for author contributions as is done
% in many journals. This is optional and at the discretion of the authors. Only add
% this information once your submission is accepted and deanonymized. 

\subsubsection*{Acknowledgments}
The authors acknowledge funding from NSF DMS-2516872.
% Use unnumbered third level headings for the acknowledgments. All
% acknowledgments, including those to funding agencies, go at the end of the paper.
% Only add this information once your submission is accepted and deanonymized. 

\bibliography{references}
\bibliographystyle{tmlr}

\newpage
\appendix
\section{Appendix}

\subsection{Simulations and implementation details}
\subsubsection{Interlaced moons toy example}
The example is constructed from 2 interlaced moons (circular arcs) in both features 1, 2 and 3. We parametrize for the 3 clusters respectively:
\begin{itemize}
    \item Cluster 1: $X_1 = 0.35 + 1.35 \sin(T_1)+\epsilon_x$ and $X_2 = 0.5+1.35\cos(T_0)+\epsilon_y$ where $T_1 \sim U(0.05\pi, 1.05\pi)$ where $\epsilon_x, \epsilon_y \sim \mathcal{N}(0,0.08^2)$. Then let $R = \sqrt{(X_2 - 0.1)^2 + (X_1 + 0.05)^2}$, $\tilde{X}_3 = -0.8\sin(1.5R)+0.2 \log(|1+X_2|) + 0.9+\epsilon_z$, $\epsilon_z \sim N(0,0.08^2)$ as well.
    \item Cluster 2: $X_1 = - 0.05 + \sin(T_2)+\epsilon_x$ and $X_2 = -0.05+\cos(T_1)+\epsilon_y$ where $T_2\sim U(0.95\pi, 1.95\pi)$ where $\epsilon_x, \epsilon_y \sim \mathcal{N}(0,0.08^2)$. Then let $R = \sqrt{(X_2 - 0.1)^2 + (X_1 + 0.05)^2}$, $\tilde{X}_3 = 0.8\sin(1.5R)+0.2 \log(|1+X_2|) - 0.7+\epsilon_z$, $\epsilon_z \sim N(0,0.08^2)$ as well.
    \item Cluster 3: $X_1 = -0.05 + 0.7\sin(T_3) + \epsilon_x$ and $X_2=0.55+0.7\cos(T_3) + \epsilon_y$ with $T_3 \sim U(-0.15\pi, 1.15\pi)$. Then let $R = \sqrt{(X_2 - 0.1)^2 + (X_1 + 0.05)^2}$, $\tilde{X}_3 = 0.8\sin(1.5R)+0.5\log(|1+X_2|) + 0.66+\epsilon_z$, $\epsilon_z \sim N(0,0.08^2)$ as well.
\end{itemize}
By construction, $\tilde{X}_3$ is a nonlinear transformation of $X_1$ and $X_2$ and is particularly  correlated with $X_2$. We decorrelate $\tilde{X}_3$ from $X_2$ to get the final $X_3$ feature: $X_3=\tilde{X}_3 - \frac{Cov(X_2, \tilde{X}_3}{Var(X_2)} X_2$. The data is then augmented with 2 features $X_4 , X_5 \sim \mathcal{U}([-1, 2])$ independently. The Cluster LOCO-Split feature importance score we obtained for this toy example in figure \ref{fig:fig1} used \texttt{sklearn}'s random forest classifier with default hyperparameters which returned a corresponding adjusted rand-index for a base spectral clustering algorithm of 0.99.

\subsubsection{Comparative simulations}
For each simulation setting, we generated labeled data $(X_i, Y_i)_{i=1}^N$ where $Y_i \in \{1, \ldots, K\}$ denotes the cluster label and each cluster $k$ contains $n_k$ observations. The data generating process first produces a low-dimensional signal representation in latent dimension $\tilde X \in \R^{d_0}$ and is embedded into higher-dimensional signal feature space $\R^{d}$. We then append pure noise features, so $X_i \in \R^{M}$ where $M=d+d_{\text{noise}}$. 

\paragraph{Gaussian Mixture.}
The standard Gaussian mixture model outputs generates the signal latent $\tilde X_i \in \R^{d_0}, \ Y_i \in \{1, \dots, K\}$. Let $\alpha \in \R^{>0}$ be a parameter controlling cluster separation, $\mu_k \in \R^{d_0}, \Sigma_k \in \R^{d_0 \times d_0}$, then observations for cluster $k$ are sampled from $\tilde X_i | Y_i = k \sim N(\alpha \mu_k, \Sigma_k)$ where $\Sigma_k$ is obtained via an \textit{onion covariance structure} \cite{qiu_generation_2006} that builds a correlation matrix layer-by-layer, hence its name, and $\mu_k$'s coordinates are sampled uniformly at random between $[-1, 1]$.

\paragraph{Gamma Mixture.}
The Gamma mixture introduces non-Gaussian marginal distributions while allowing dependence among features through a Gaussian copula. For each cluster $k$, let $R_k$ denote a cluster-specific correlation matrix and let $F_{kj}$ be the CDF of a gamma distribution with shape parameter $a_j$ and cluster-specific scale parameter $s_{kj}$, $\Gamma(a_j, s_{kj})$ . We define the joint distribution of $\tilde X_i \mid Y_i=k$ by the Gaussian copula
\[C_{R_k}(u_1,\ldots,u_{d_0})=\Phi_{R_k}\left(\Phi^{-1}(u_1),\ldots,\Phi^{-1}(u_{d_0})\right),
\]
where $\Phi_{R_k}$ is the CDF of a  $\mathcal{N}(0, R_k)$ with $R_k \in \R^{d_0, d_0}$, and $\Phi$ is the CDF of $\mathcal{N}(0, I)$. The samples are generated by drawing $z_i \sim \mathcal{N}(0,R_k)$, setting $U_{ij}=\Phi(z_{ij})$, and then applying the inverse gamma CDF $\tilde X_{ij}=F_{kj}^{-1}(U_{ij})$. This construction gives each coordinate a gamma marginal distribution while using $R_k$ to control the dependence structure within cluster $k$.

\paragraph{Interlaced Moons/Circles.} We generate half moons and circles in a two dimensional lower dimension first using mixtures of circles and half-moons. For each cluster $k$, a radius $r_k$ and center $c_k \in \mathbb{R}^2$ are sampled, and the cluster shape is chosen to be either a full circle or a half-circle from a given probability vector. Points are sampled along the corresponding circular arc with additive Gaussian noise:
$X_i = c_k + r_k(\cos \theta_i,\sin \theta_i) + \varepsilon_i$,
where $\theta_i$ is sampled either on $[0,2\pi]$ for circles or on an interval of length $\pi$ for half-moons, with a random orientation. Cluster centers are placed so that the pairwise overlap between the underlying circles is bounded by a fixed maximum overlap percentage. This produces nonlinear, partially interlaced cluster structures. 

\paragraph{Embedding data in higher dimensions and noise features}
For high-dimensional simulations, we embed the low-dimensional signal into a larger ambient space using random feature maps: we generate random orthonormal projection maps, bringing the $\tilde \mX \in \R^{N \times d_0}$ latent observations into $\hat P \tilde \mX = \mX \in \mathbb{R}^{N \times d}$. We also construct cluster-specific embeddings, allowing different clusters to be transformed through different random maps. Finally, we add pure noise signal $\textbf{S} \in \R^{N\times d_{\text{noise}}}$ sampled from distributions such as Gaussian, Gamma with $a=1, s=1$, Student-$t$, uniform, triangular, or Laplace distributions. The final data generated by our simulator becomes $\mX = [\hat \mX, \textbf{S}] \in \mathbb{R}^{N \times M}$.

\paragraph{Implementation details} To obtain the results shown in Figures~\ref{fig:fig2} and \ref{fig:supp2a}--\ref{fig:supp2c}, we implemented Cluster LOCO-MP and corresponding RAMPART with \texttt{sklearn}'s decision tree classifier with default hyperparameters for the Gaussian mixtures and the random forest classifier with also default hyperparameters in the cases of the interlaced moons/circles and Gamma mixture. Additionally, the clustering solution is not held fixed for the evaluated methods, only a \textit{base clustering model} is fixed when possible: for example, permutation feature importance needs refitting multiple times with permuted features - the clustering algorithm is therefore refitted and not held fixed, for the neuralized KMeans, we use the original architecture which does not fit the Lloyd's KMeans, but an equivalent objective optimized via a neural network, and finally for both our scores, the clustering algorithm is re-fitted on splits, or on minipatches. Unless otherwise stated, replicates within each simulation setting use the same generated dataset with different random seeds for data splitting and model fitting. Thus, the reported variability reflects randomness in data splitting, minipatch sampling, and stochastic model fitting, as applicable, rather than variation across independently generated datasets.

\subsection{Cluster LOCO-RAMPART algorithm}

\begin{algorithm}[htbp]
    \caption{Cluster LOCO-RAMPART}
    \label{alg:rampart}
    \begin{algorithmic}[1]
    \Require Unlabeled data $\mX\in\R^{N\times M}$; top-$k$ target $1\leq k<M$; minipatch sizes $(n,m)$ with $1\leq n<N$, $m\geq1$;
    minipatches per round $B$.

    \State Set $t=0$. Initialize active feature set $S_0=[M]$.

    \While{$|S_t|>k$}
        \State Set the feature minipatch size $m_t=\min\{m,|S_t|-1\}$.
        \State Obtain round-$t$ Cluster LOCO-MP estimates:
        \[ \{\hat\Delta_j^{(t)}\}_{j\in S_t} \leftarrow \text{Cluster LOCO-MP}(\mX_{:,S_t},n,m_t,B) \]
        retaining the original feature indices. If any score is undefined, stop and report insufficient valid predictions.
        \State Rank features in descending order:
        \[ \hat\Delta_{\hat\tau_1^t}^{(t)} \geq \cdots \geq \hat\Delta_{\hat\tau_{|S_t|}^t}^{(t)}\]
        breaking ties by original feature index.
        \State Retain the top half, or $k$ features if larger:
        \[q_t=\max\{k,\lceil |S_t|/2\rceil\}, \qquad S_{t+1}= \{\hat\tau_1^t,\ldots,\hat\tau_{q_t}^t\}\]
        \State Store the retained scores: $d_j=\hat\Delta_j^{(t)}$ for $j\in S_{t+1}$.
        \State Set $t\leftarrow t+1$.
    \EndWhile

    \Ensure Selected features $S_t$, with $|S_t|=k$, and scores $\{d_j\}_{j\in S_t}$ from the final screening round.
    \end{algorithmic}
\end{algorithm}

\newpage
\subsection{Additional figures}

\begin{figure}[h!]
    \centering
    \includegraphics[width=\linewidth]{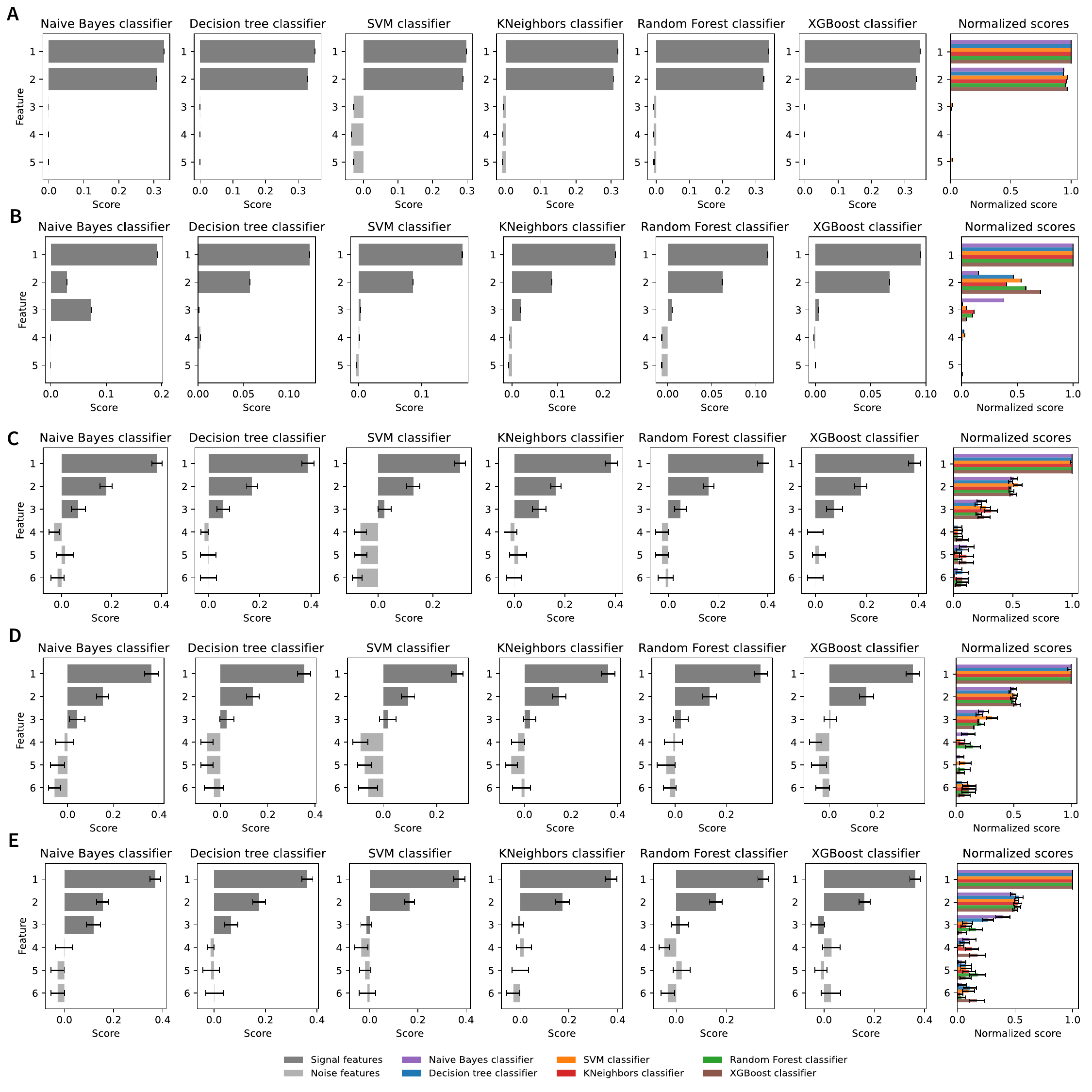}
    \caption{\textbf{A.} 3GMM example with independent mixtures. \textbf{B.} Toy example described in section 3. \textbf{C-E}, 5GMM example with correlated features, features are correlated with a Toeplitz structure, with respectively 0 (independent), 0.2 and 0.8 correlation between consecutive features. Signal features were generated on a simplex with ordered incremental importance. Results were run on 20 replicates with different seeds to incorporate classifier and clustering variability. All but \textbf{B} used KMeans clustering algorithm with ARI scores of 1 for the base clustering, \textbf{B }was run with Spectral Clustering.}
    \label{fig:supp_fig5}
\end{figure}

\begin{figure}[h!]
    \centering
    \begin{subfigure}{\linewidth}
    \includegraphics[width=\linewidth]{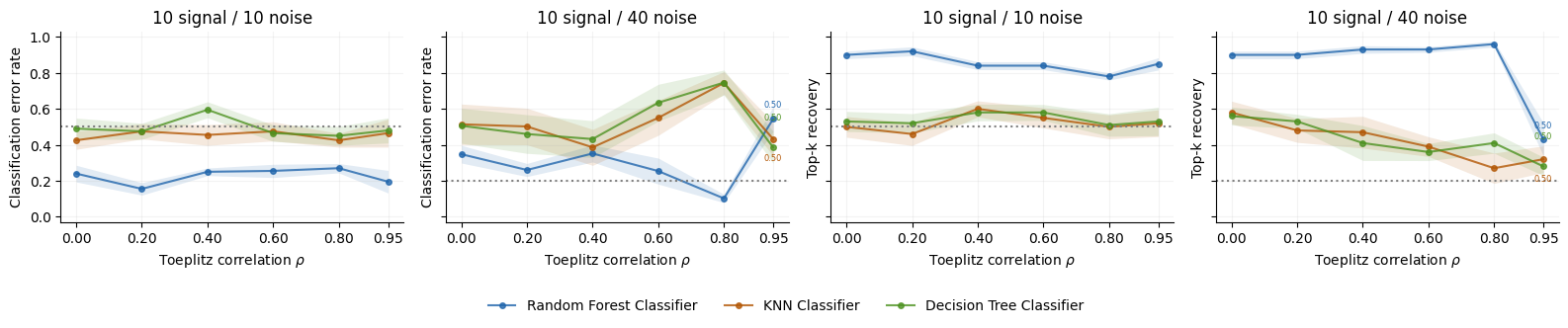}
    \caption{Cluster LOCO-Split feature classification error rate and top-$10$ feature recovery in a 5-GMM simulated example with correlations given different classifier choices.}
    \end{subfigure}
    \begin{subfigure}{\linewidth}
    \includegraphics[width=\linewidth]{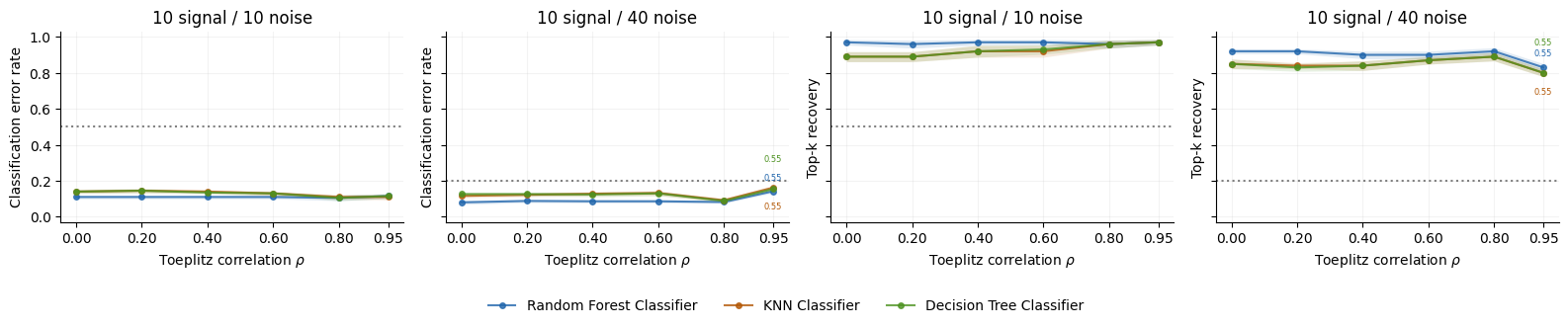}
    \caption{Cluster LOCO-MP feature classification error rates and top-$10$ feature recovery in the same 5-GMM simulated example with correlations given different classifier choices.}
    \end{subfigure}
    \caption{This simulated example was obtained with $N=500$, 10 replicates with varying seeds, KMeans base clustering algorithm and $B=500$ minipatches. Correlation structure in the signal features follows the same Toeplitz structure as the previous experiment. Average clustering ARI is omitted when $\geq 0.99$. CER reported is computed using (false positive+false negatives)/total feature. }
    \label{fig:supp_fig6}
\end{figure}

\begin{figure}[h!]
    \centering
    \includegraphics[width=\linewidth]{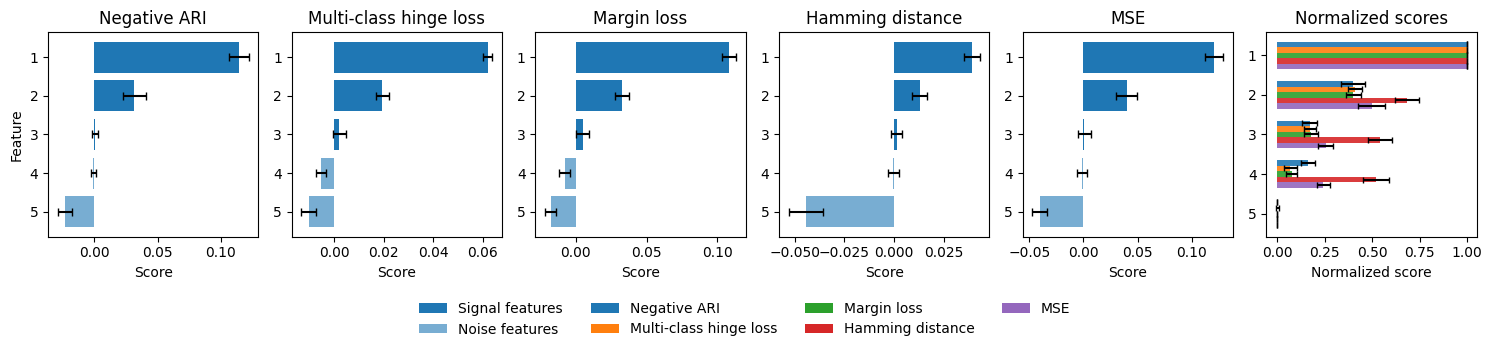}
    \caption{Comparison of Cluster LOCO-Split scores with different measures of error/dissimilarity averaged over 100 replicates with varying seeds for split-stability} for the toy example described in Section 3.1. Mean squared error and Hamming distance show the more difference in normalized scores compared to other feature importance scores but we see that this is an effect of the scaling, feature 5 has a "worse" effect for these two errors relative to the rest of the scores, normalizing inflates the contribution of feature 2, 3 and 4 since all scores are normalized to be positive with highest importance at 1.
    \label{fig:supp_fig7}
\end{figure}

\begin{figure}[!ht]
    \centering
    %\hspace*{-1.2cm}
    \begin{subfigure}{\linewidth}
    \includegraphics[width=\linewidth]{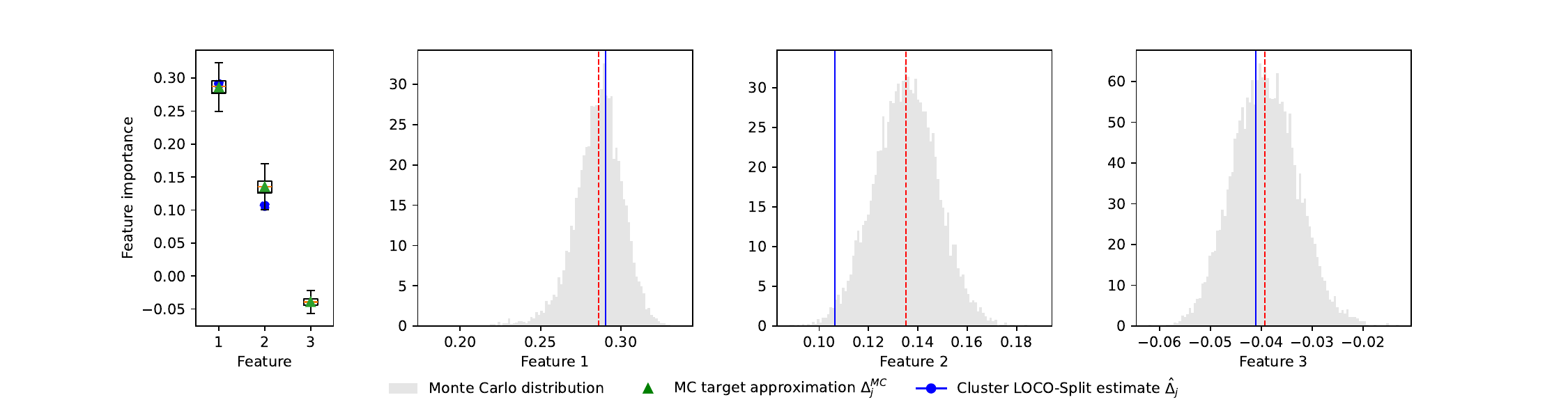}
    \caption{Cluster LOCO-Split population target approximation in a 3 Gaussian Mixture model with 2 signal feature and 1 standard Gaussian noise feature. Base algorithm used was KMeans with random forest classifier.  The training dataset, its cluster assignments, and the fitted full-feature and reduced-feature transfer classifiers are held fixed. Each of $R_{\mathrm{mc}}=5000$ independent repetitions generates a fresh calibration dataset of the same size as the training dataset.  Monte Carlo standard errors are computed as the between-repetition standard deviation divided by $\sqrt{R_{\mathrm{mc}}}$. }
    \label{fig:supp_fig8a}
    \end{subfigure}
    \begin{subfigure}{\linewidth}
    \hspace*{-2cm}
    \includegraphics[width=1.2\linewidth]{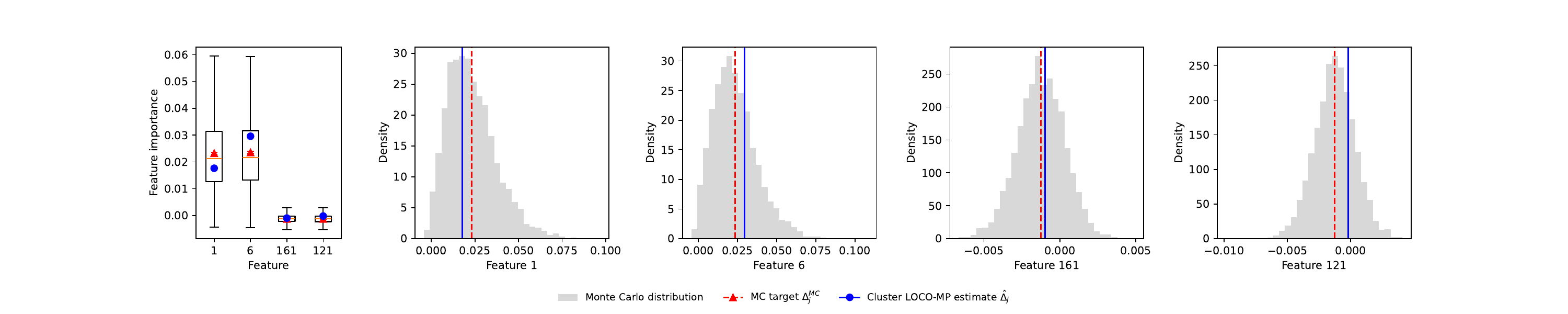}
    \caption{Cluster LOCO-MP population target approximation in a 10 signal features, 190 noise features 5-GMM example with 5000 observations. Base clustering algorithm used was KMeans with random forest classifier. Shown are two signal features 1 and 6 and two noise features 161 and 121 chosen at random. Each of $R_{\mathrm{mc}}$ independent repetitions generates a fresh dataset and performs a complete Cluster LOCO-MP fit using $B_{\mathrm{mp}}=5000$ minipatches with all hyperparameters fixed. Monte Carlo standard errors are computed as the between-repetition standard deviation divided by $\sqrt{R_{\mathrm{mc}}=5000}$. Unlike the conditional Split target, this target averages over fresh datasets and complete fits.}
    \label{fig:supp_fig8b}
    \end{subfigure}
    \caption{\small \textbf{Monte Carlo approximation of the population target} for Cluster LOCO-Split and Cluster LOCO-MP feature importance scores.} 
    
\end{figure}

\begin{figure}[htbp]
    \centering

    \begin{subfigure}[t]{0.325\textwidth}
        \vspace{0pt}
        \centering
        \begin{tcolorbox}[simbox]
            \centering
            % \vspace{1em}

            \vspace{1em}
            \panelimg{A. Gaussian Mixture}{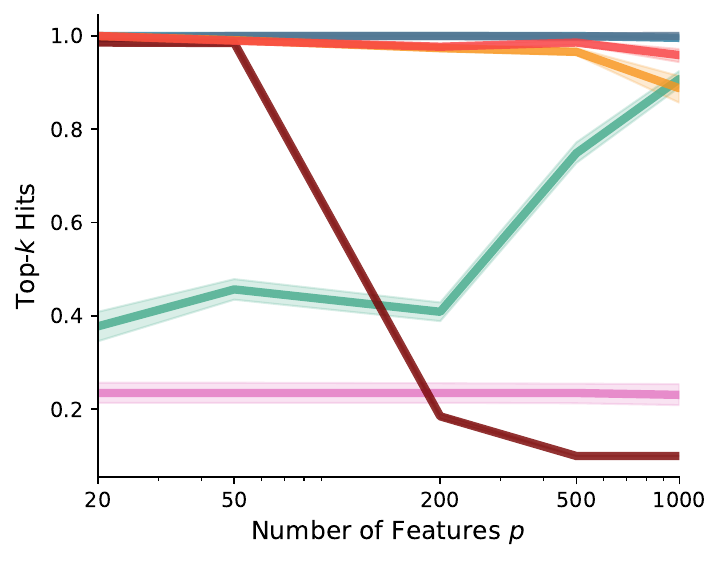}

            % \vspace{1em}
            % \panelimg{1c.}{Figures/gaussian_harder.pdf}

            \vspace{1em}
            \panelimg{}{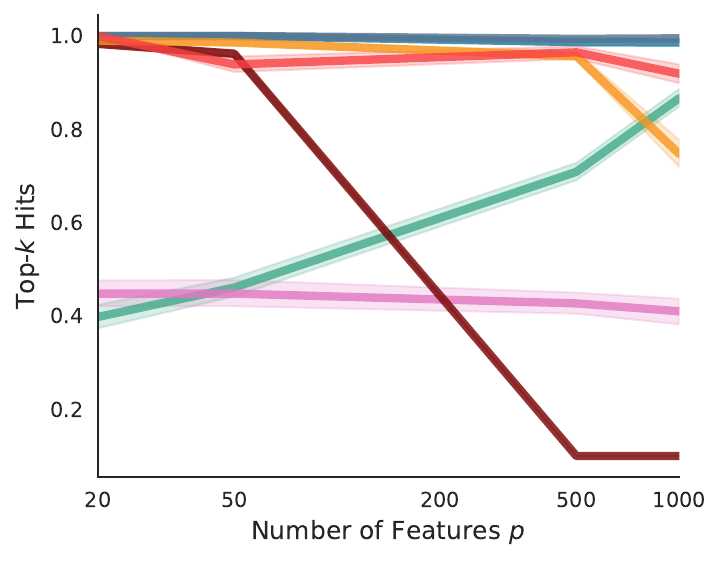}
        \end{tcolorbox}
    \end{subfigure}
    \hfill
    \begin{subfigure}[t]{0.325\textwidth}
        \vspace{0pt}
        \centering
        \begin{tcolorbox}[simbox]
            \centering
            % \vspace{1em}
            % % \panelimg{2a.}{Figures/gamma_easier.pdf}

            \vspace{1em}
            \panelimg{B. Gamma Mixture}{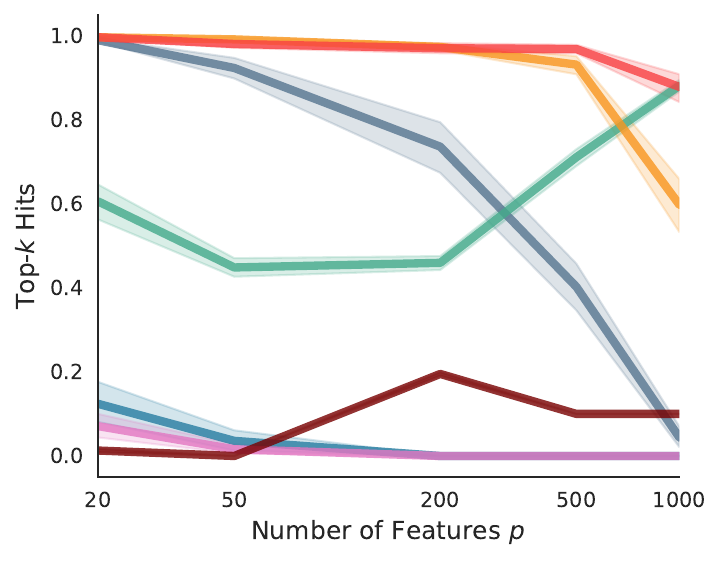}

            % \vspace{1em}
            % \panelimg{2c.}{Figures/gamma_harder.pdf}

            \vspace{1em}
            \panelimg{}{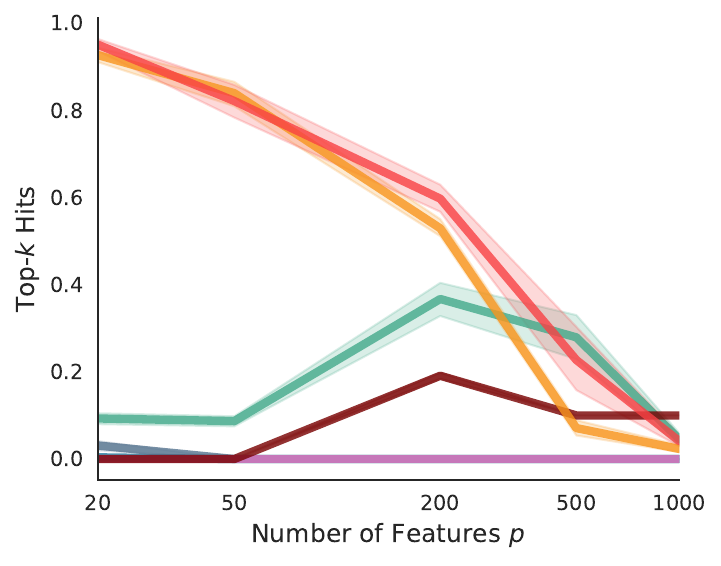}
        \end{tcolorbox}
    \end{subfigure}
    \hfill
    \begin{subfigure}[t]{0.325\textwidth}
        \vspace{0pt}
        \centering
        \begin{tcolorbox}[simbox]
            \centering
            % \vspace{1em}
            % \panelimg{3a.}{Figures/moons_easier.pdf}

            \vspace{1em}
            \panelimg{C. Moons/Circles }{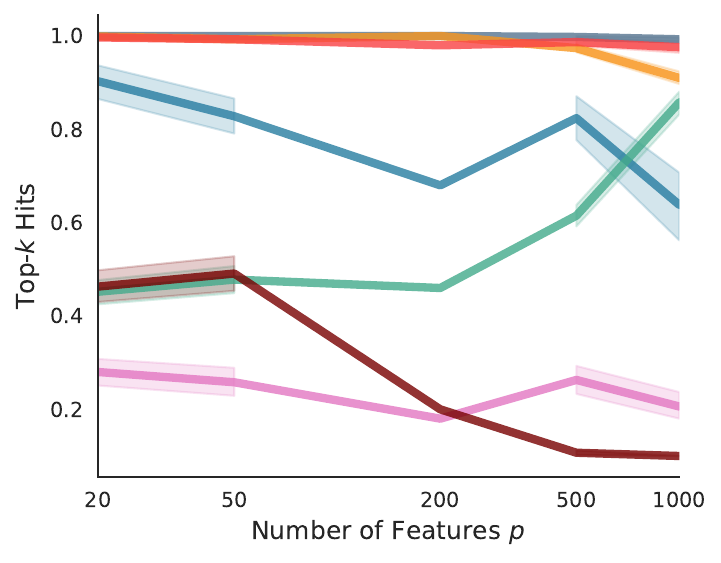}

            % \vspace{1.3em}
            % \panelimg{3c.}{Figures/moons_harder.pdf}

            \vspace{1 em}
            \panelimg{}{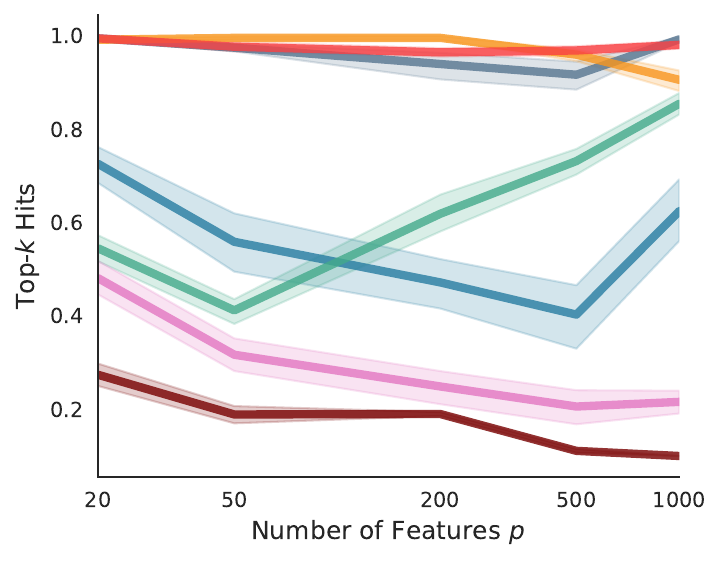}
        \end{tcolorbox}
    \end{subfigure}

    \vspace{0.5em}

    \begin{subfigure}[t]{\textwidth}
        \centering
        \includegraphics[width=\linewidth]{Figures/fig2_legend.pdf}
    \end{subfigure}

    \caption{\small \textbf{Top-$k$ hits} in three simulation settings for signal features $p^* = 10$ and noise features features $p_{\text{noise}} \in \{10, 40, 190, 490, 990\}$ with $K=3$ clusters with $N=500$ observations per cluster. Cluster LOCO-MP was run with $\alpha_M = \alpha_N = 0.2$ and $B=5000$, Cluster LOCO-RAMPART with $B_{\text{rampart}} = 1000$. Model agnostic methods are obtained with KMeans for Gaussian mixtures, Gamma mixture EM for Gamma mixture, Spectral Clustering for Moons and Circles. Model-specific methods are reported with their base model in the legend. Top-$k$ hits were reported averaged over 100 replicates.}
    \label{fig:supp2a}
\end{figure}

\begin{figure}[htbp]
    \centering

    \begin{subfigure}[t]{0.325\textwidth}
        \vspace{0pt}
        \centering
        \begin{tcolorbox}[simbox]
            \centering
            % \vspace{1em}

            \vspace{1em}
            \panelimg{A. Gaussian Mixture}{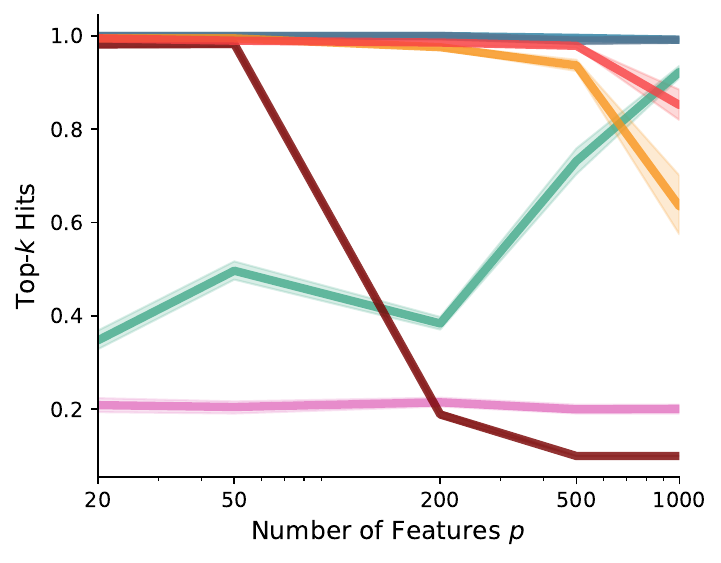}

            % \vspace{1em}
            % \panelimg{1c.}{Figures/gaussian_harder.pdf}

            \vspace{1em}
            \panelimg{}{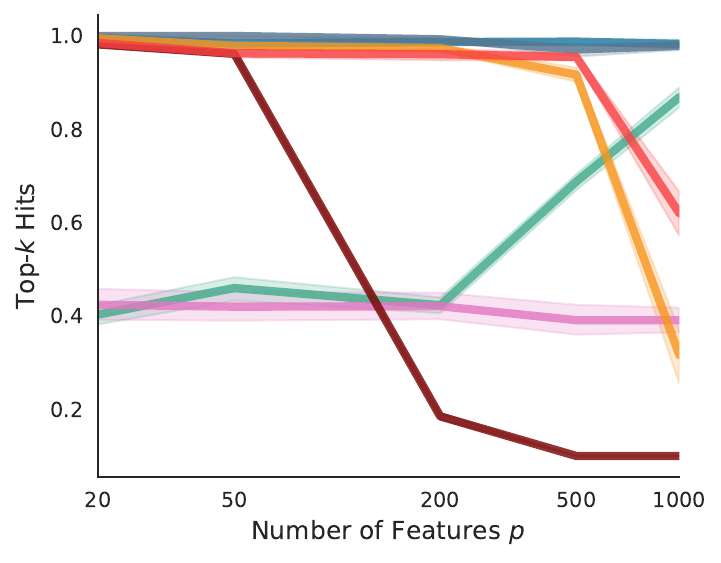}
        \end{tcolorbox}
    \end{subfigure}
    \hfill
    \begin{subfigure}[t]{0.325\textwidth}
        \vspace{0pt}
        \centering
        \begin{tcolorbox}[simbox]
            \centering
            % \vspace{1em}
            % % \panelimg{2a.}{Figures/gamma_easier.pdf}

            \vspace{1em}
            \panelimg{B. Gamma Mixture}{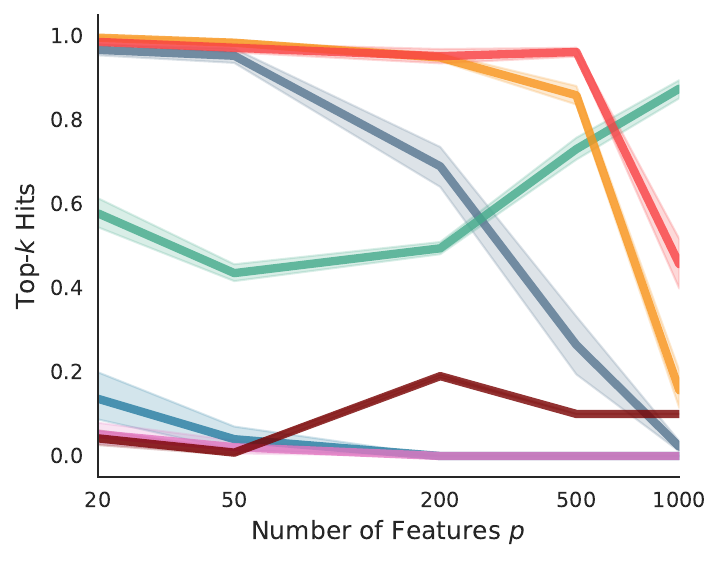}

            % \vspace{1em}
            % \panelimg{2c.}{Figures/gamma_harder.pdf}

            \vspace{1em}
            \panelimg{}{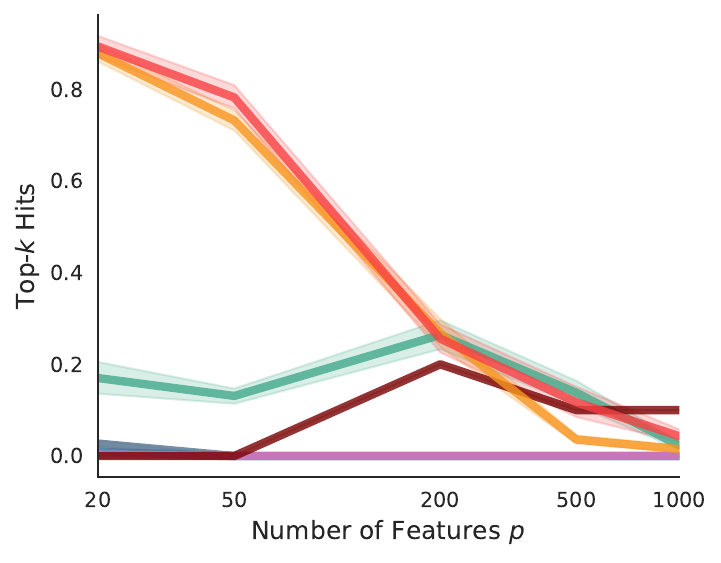}
        \end{tcolorbox}
    \end{subfigure}
    \hfill
    \begin{subfigure}[t]{0.325\textwidth}
        \vspace{0pt}
        \centering
        \begin{tcolorbox}[simbox]
            \centering
            % \vspace{1em}
            % \panelimg{3a.}{Figures/moons_easier.pdf}

            \vspace{1em}
            \panelimg{C. Moons/Circles }{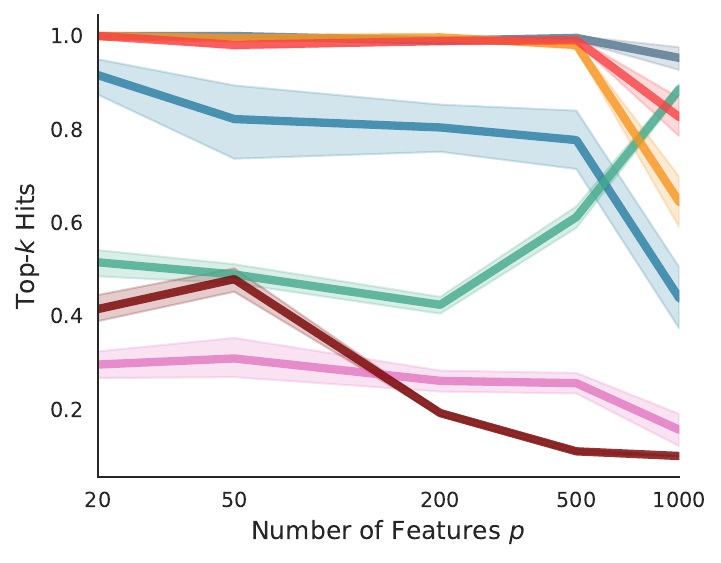}

            % \vspace{1.3em}
            % \panelimg{3c.}{Figures/moons_harder.pdf}

            \vspace{1 em}
            \panelimg{}{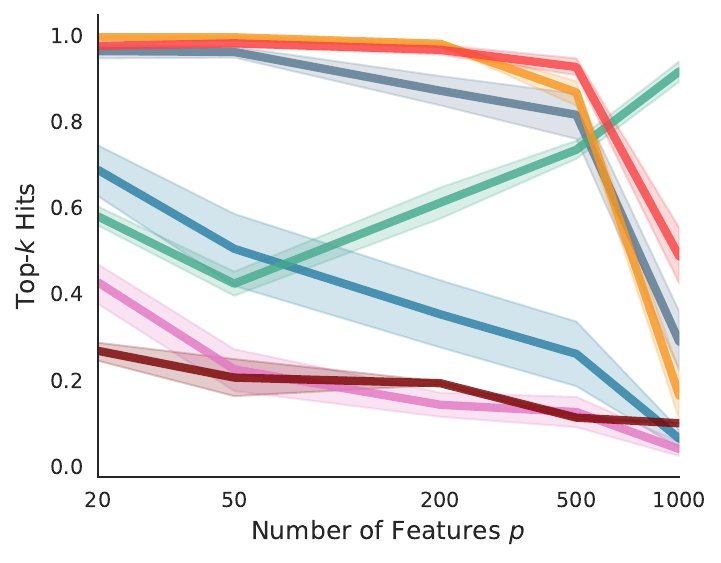}
        \end{tcolorbox}
    \end{subfigure}

    \vspace{0.5em}

    \begin{subfigure}[t]{\textwidth}
        \centering
        \includegraphics[width=\linewidth]{Figures/fig2_legend.pdf}
    \end{subfigure}

    \caption{\small \textbf{Top-$k$ hits} in three simulation settings for signal features $p^* = 10$ and noise features features $p_{\text{noise}} \in \{10, 40, 190, 490, 990\}$ with $K=3$ clusters with $N=300$ observations per cluster. Cluster LOCO-MP was run with $B=5000$, Cluster LOCO-RAMPART with $B_{\text{rampart}} = 1000$ and adaptive $\alpha_N \in (0.1, 0.5)$, $\alpha_p \in (0.1, 0.5)$. Model agnostic methods are obtained with KMeans for Gaussian mixtures, Gamma mixture EM for Gamma mixture, Spectral Clustering for Moons and Circles. Model-specific methods are reported with their base model in the legend. Top-$k$ hits were reported averaged over 100 replicates.}
    \label{fig:supp2b}
\end{figure}

\begin{figure}[htbp]
    \centering

    \begin{subfigure}[t]{0.325\textwidth}
        \vspace{0pt}
        \centering
        \begin{tcolorbox}[simbox]
            \centering
            % \vspace{1em}

            \vspace{1em}
            \panelimg{A. Gaussian Mixture}{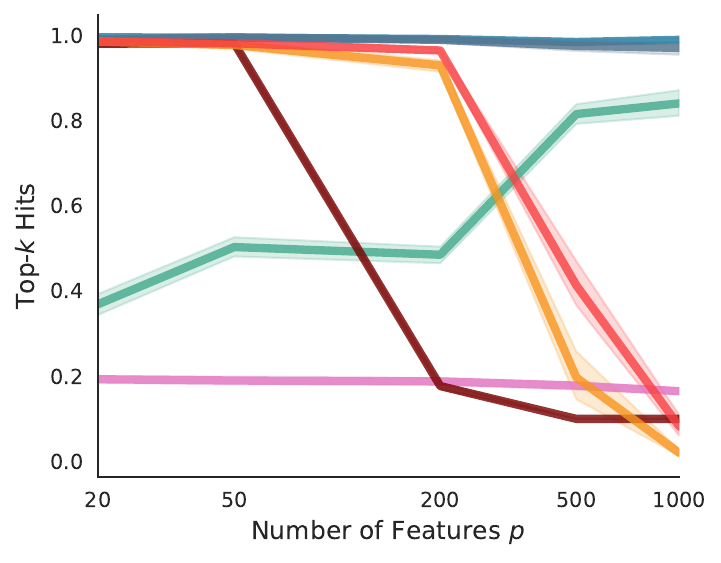}

            % \vspace{1em}
            % \panelimg{1c.}{Figures/gaussian_harder.pdf}

            \vspace{1em}
            \panelimg{}{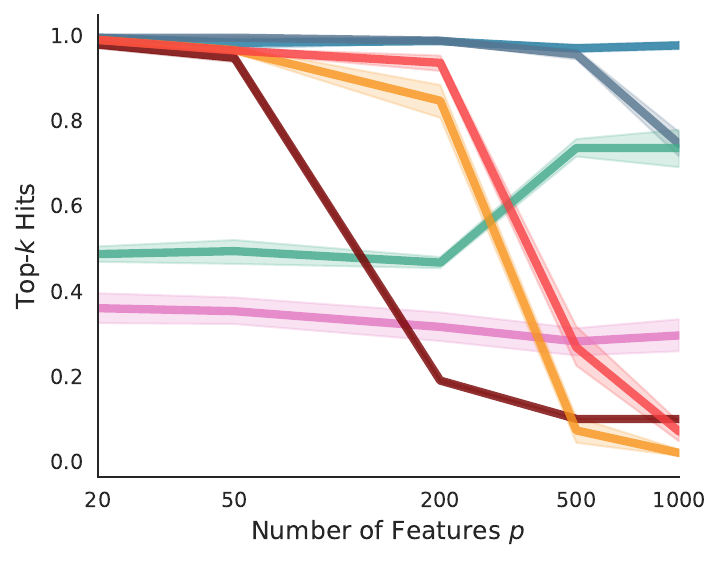}
        \end{tcolorbox}
    \end{subfigure}
    \hfill
    \begin{subfigure}[t]{0.325\textwidth}
        \vspace{0pt}
        \centering
        \begin{tcolorbox}[simbox]
            \centering
            % \vspace{1em}
            % % \panelimg{2a.}{Figures/gamma_easier.pdf}

            \vspace{1em}
            \panelimg{B. Gamma Mixture}{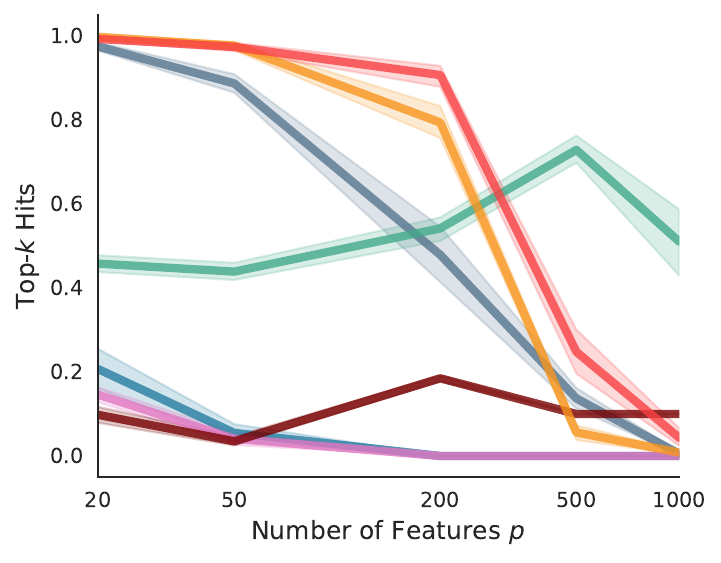}

            % \vspace{1em}
            % \panelimg{2c.}{Figures/gamma_harder.pdf}

            \vspace{1em}
            \panelimg{}{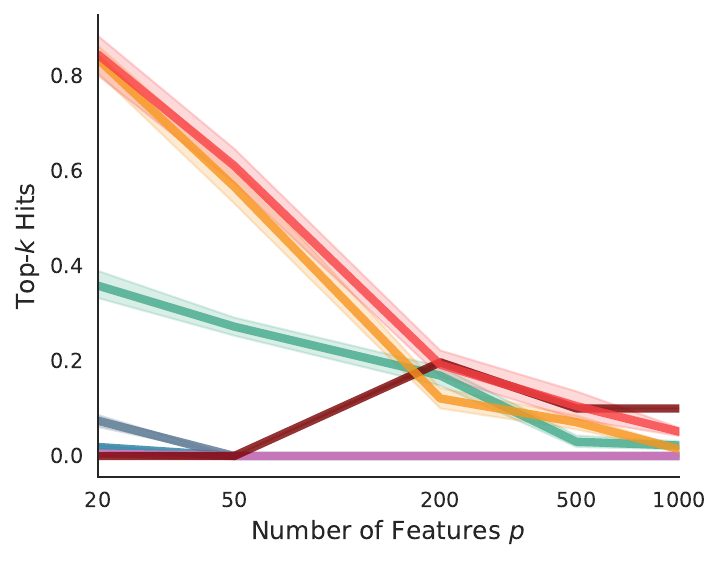}
        \end{tcolorbox}
    \end{subfigure}
    \hfill
    \begin{subfigure}[t]{0.325\textwidth}
        \vspace{0pt}
        \centering
        \begin{tcolorbox}[simbox]
            \centering
            % \vspace{1em}
            % \panelimg{3a.}{Figures/moons_easier.pdf}

            \vspace{1em}
            \panelimg{C. Moons/Circles }{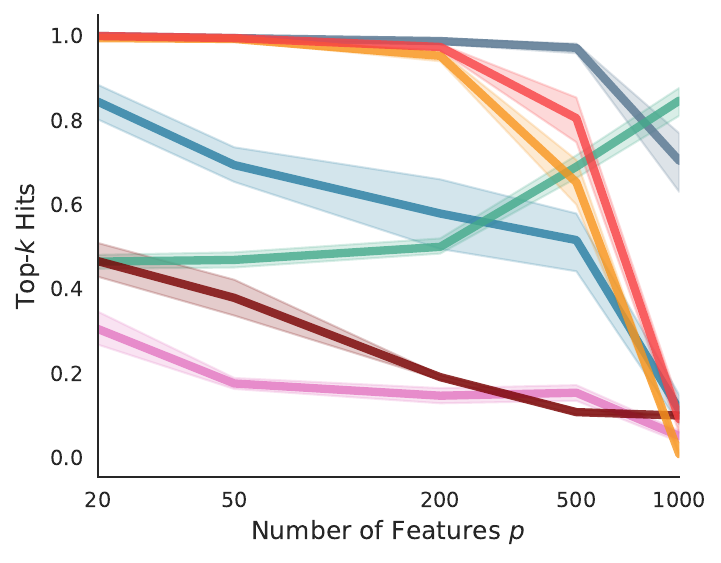}

            % \vspace{1.3em}
            % \panelimg{3c.}{Figures/moons_harder.pdf}

            \vspace{1 em}
            \panelimg{}{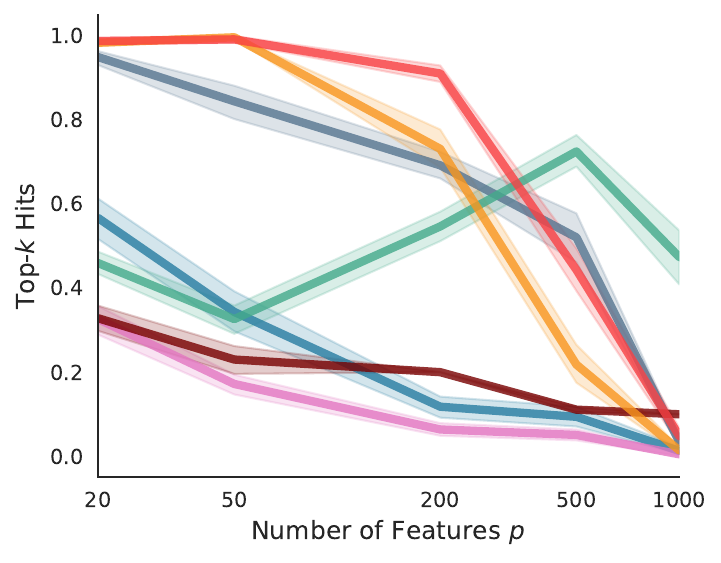}
        \end{tcolorbox}
    \end{subfigure}

    \vspace{0.5em}

    \begin{subfigure}[t]{\textwidth}
        \centering
        \includegraphics[width=\linewidth]{Figures/fig2_legend.pdf}
    \end{subfigure}

    \caption{\small \textbf{Top-$k$ hits} in three simulation settings for signal features $p^* = 10$ and noise features features $p_{\text{noise}} \in \{10, 40, 190, 490, 990\}$ with $K=3$ clusters with $N=100$ observations per cluster. Cluster LOCO-MP was run with $\alpha_M = \alpha_N = 0.2$ and $B=5000$, Cluster LOCO-RAMPART with $B_{\text{rampart}} = 1000$. Model agnostic methods are obtained with KMeans for Gaussian mixtures, Gamma mixture EM for Gamma mixture, Spectral Clustering for Moons and Circles. Model-specific methods are reported with their base model in the legend. Top-$k$ hits were reported averaged over 100 replicates.}
    \label{fig:supp2c}
\end{figure}

\begin{figure}[!ht]
    \centering
    \includegraphics[width=\linewidth]{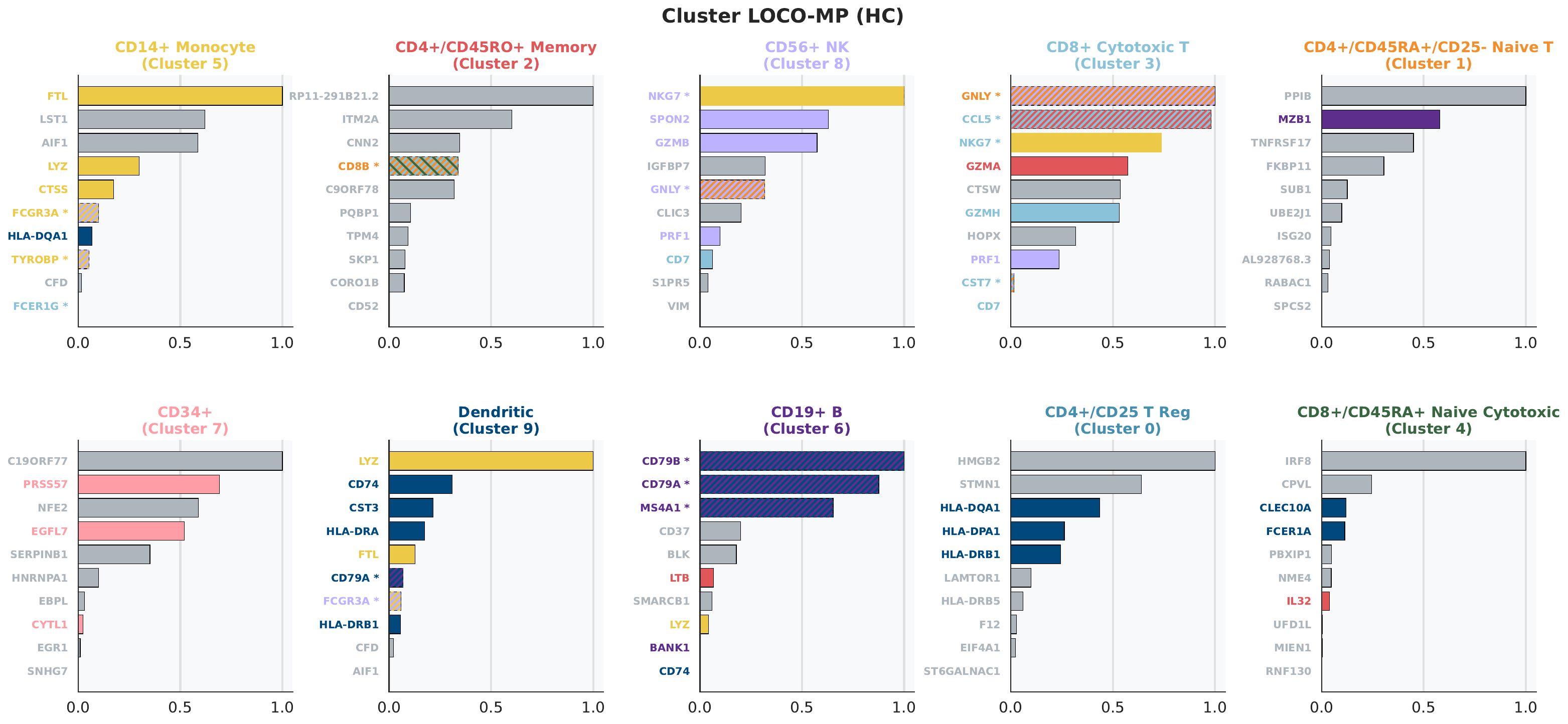}
    \caption{\small \textbf{Top 10 most important genes for clusters via Cluster LOCO-MP} feature importance scores: Hierarchical Clustering was used as base clustering algorithm. Labels were aligned with the reported \textit{purified} labels. Known markers from our reference set are colored with the cell-type they identify.}
    \label{fig:supp3a}
\end{figure}

\begin{figure}[!ht]
    \centering
    \includegraphics[width=\linewidth]{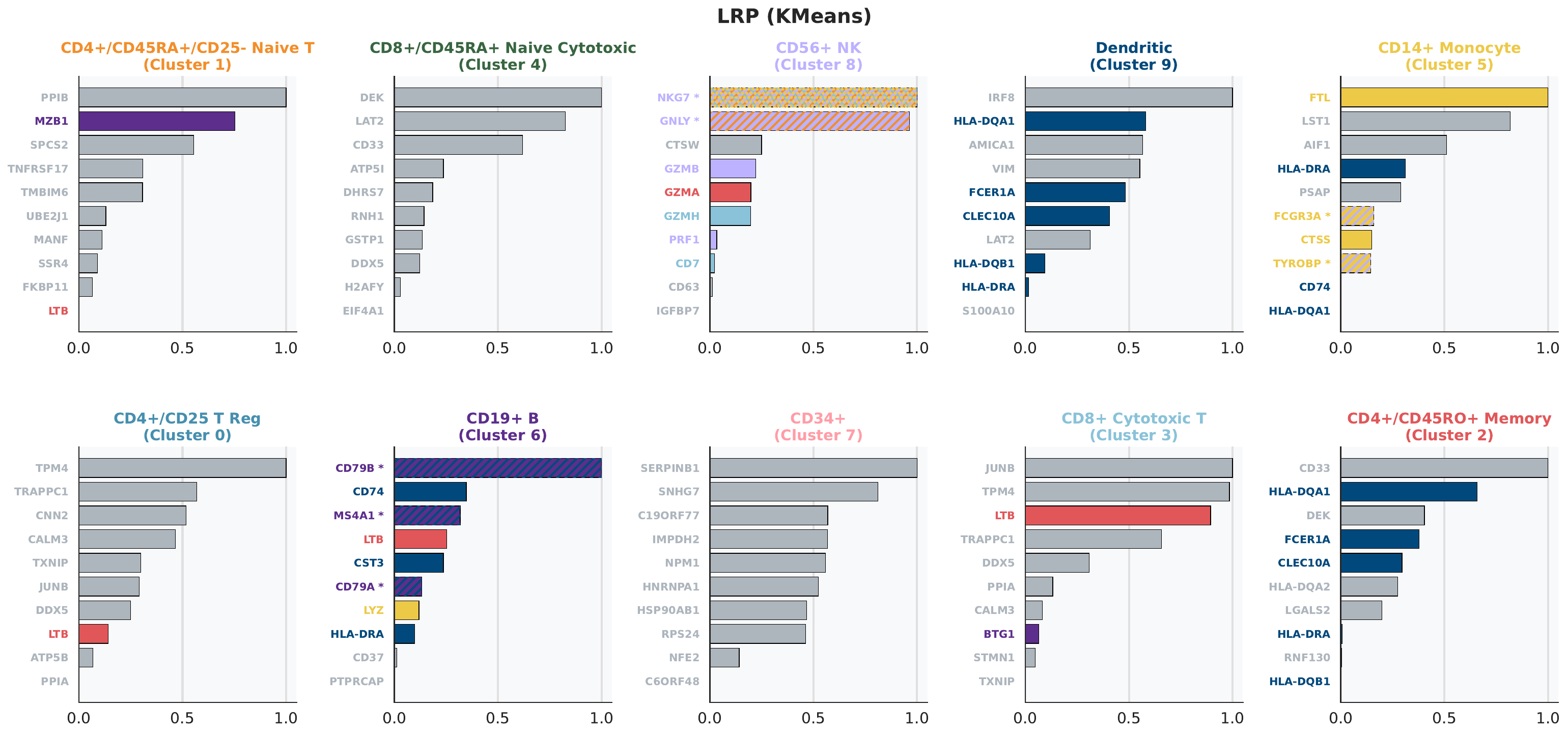}
    \caption{\small \textbf{Top 10 most important genes for clusters via LRP} feature importance scores, LRP uses a neuralized KMeans clustering algorithm. Labels were aligned with the reported \textit{purified} labels. Known markers from our reference set are colored with the cell-type they identify.}
    \label{fig:supp3b}
\end{figure}

\begin{figure}[!ht]
    \centering
    \hspace*{-1.2cm}
    \includegraphics[width=\linewidth]{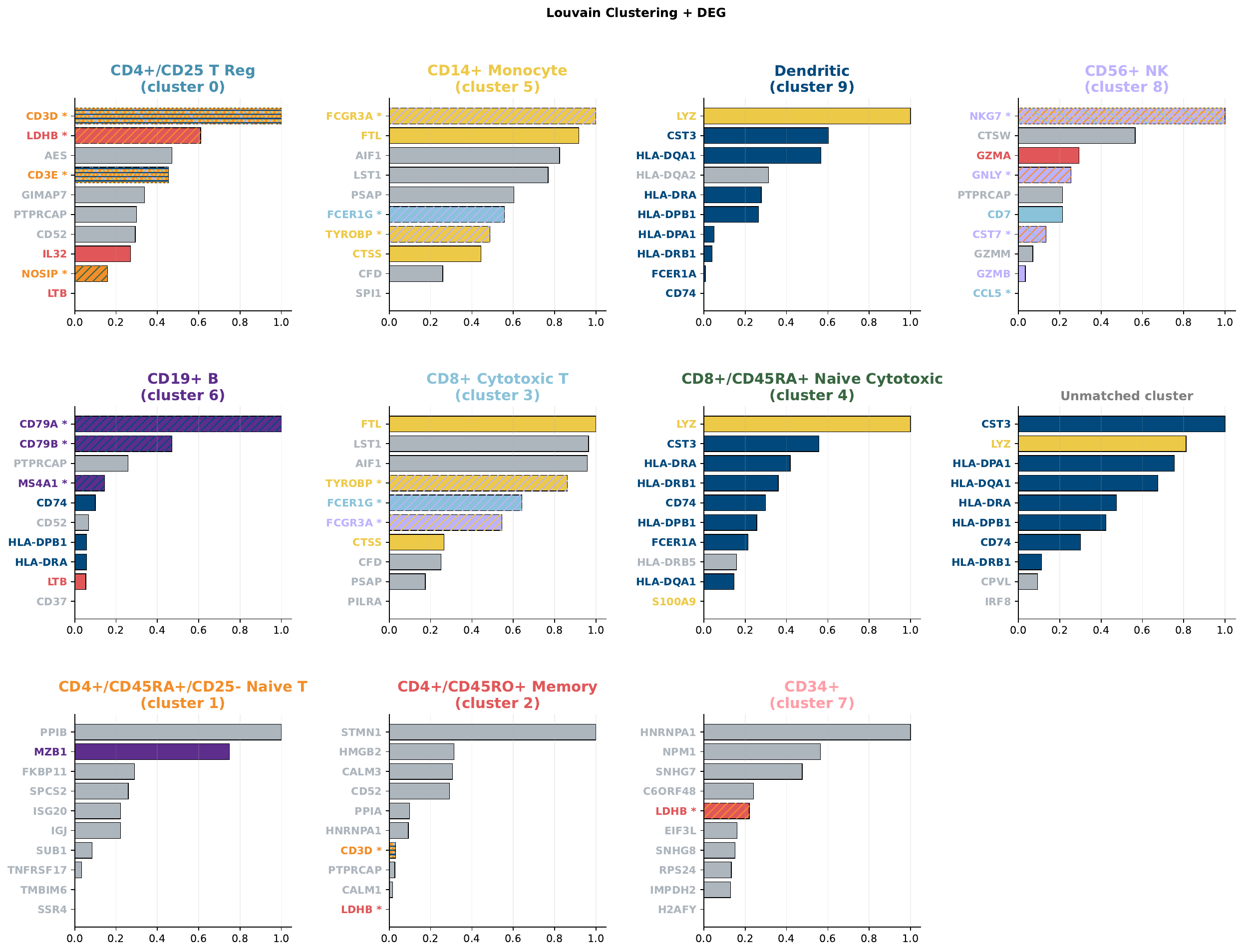}
    \caption{\small \textbf{Top 10 most important DEGs} obtained with Louvain clustering and t-test with BH correction. DEGs were filtered at significance level of 0.05 and ranked by z-score magnitude. Labels were aligned with the reported \textit{purified} labels. Known markers from our reference set are colored with the cell-type they identify. We see the effect of overclustering in cluster 9, cluster 4 and the unmatched cluster.}
    \label{fig:supp3c}
\end{figure}

\begin{figure}[!ht]
    \centering
    % \hspace*{-1.2cm}
    \includegraphics[width=
    \linewidth]{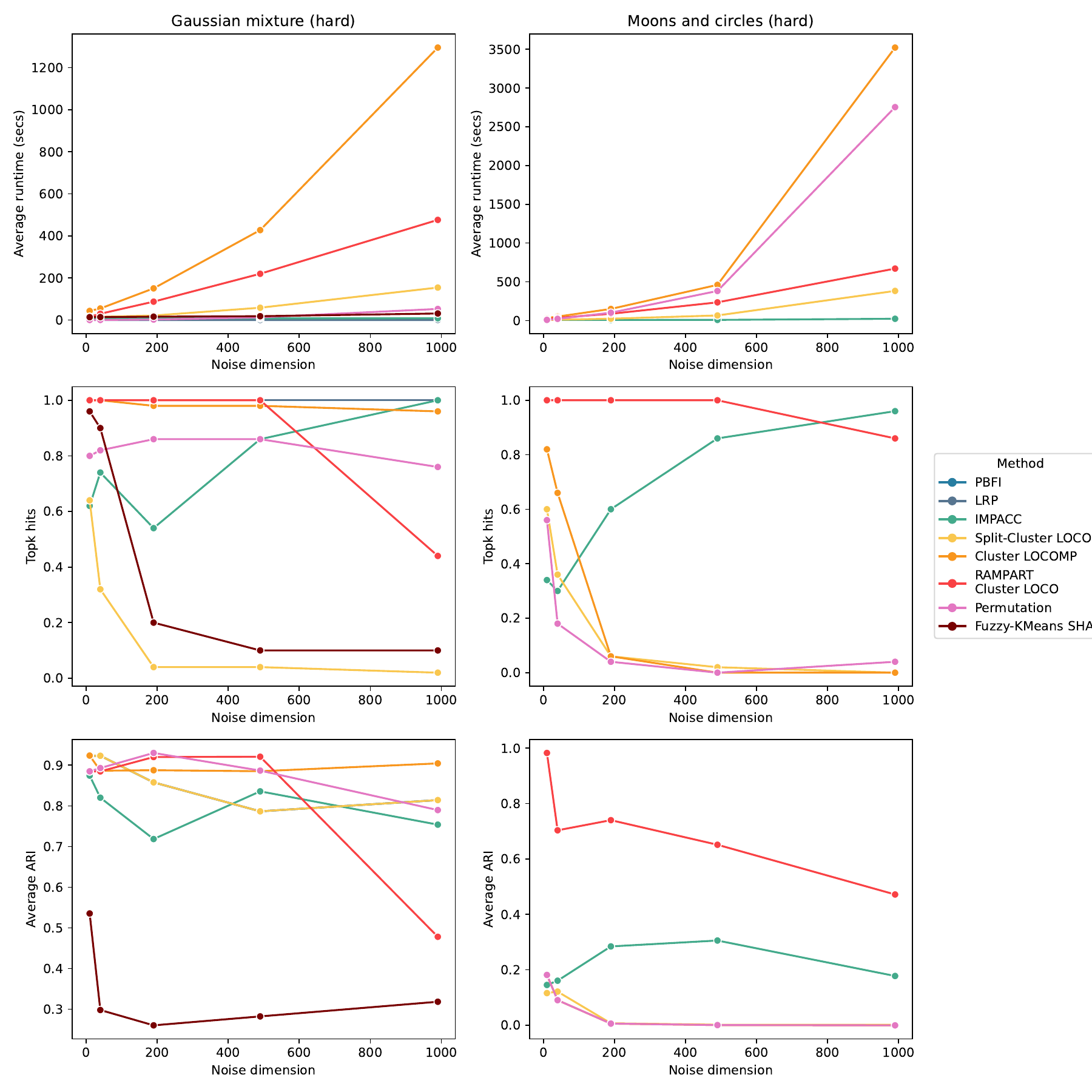}
    \caption{\textbf{Computational costs, feature recovery performance and ARI } across methods for hard Gaussian mixtures and moons and circles simulations. Cluster LOCO RAMPART is especially efficient at balancing performance and computational costs when using a more costly base clustering algorithm (Spectral Clustering). However, the performances are not exactly comparable, since methods were run with default recommended budget.}
    \label{fig:supp_fig15}
\end{figure}

\clearpage
\newpage
\subsection{Appendix tables}

\noindent As detailed in the simulation section, we ran FCM SHAP with 20 iterations as recommended by typical SHAP methods, while this allows for fast computation, it sacrifices quality of the approximation as shown in the top-$k$ recovery results. Feature permutation method was also run with a fixed permutation iteration budget of 10 iterations per feature. The number of minipatch should be set high and budgeted in advance, and for these settings, we find that $B=5000$ is enough and report the corresponding runtime, the budget for RAMPART was set to $B=1000$ minipatches. IMPACC uses a data adaptive strategy to allocate resources according to a fixed budget which was set to the default 300 iterations.

\begin{table}[htbp]
\centering
\caption{Computational cost (in seconds) for hard Gaussian mixture simulation (Figure \ref{fig:fig2}A)}
\label{tab:runtime-gaussian-additional}
\resizebox{\textwidth}{!}{%
\begin{tabular}{l|cccccccc}
\toprule
Simulation & PBFI & LRP & IMPACC & Cluster LOCO-Split & Cluster LOCO-MP & CL-RAMPART & Permutation & FCM SHAP \\
\midrule
Gaussian ($n=200, \ p=20$)
& \textbf{0.0039 (0.00)}
&  0.5675 (0.23)
& 7.6764 (0.20)
& 0.4631 (0.38)
& 4.9203 (0.01)
& 2.2297 (0.04)
& 0.0944 (0.02)
& 3.1780 (0.27) \\
Gaussian ($n=200, \ p=1{,}000$)
& \textbf{0.0882 (0.00)}
& 0.6436 (0.21)
& 7.9988 (0.10)
& 85.0434 (2.73)
& 39.0535 (0.34)
& 13.1640 (0.14)
& 53.1354 (0.47)
& 7.1672 (0.27) \\
\bottomrule
\end{tabular}%
}
\end{table}

\begin{table}[htbp]
\centering
\caption{Computational cost (in seconds) for hard Gamma mixture and Moons and circles simulations (Figure \ref{fig:fig2}B and C)}
\label{tab:runtime-gamma-moon}
\resizebox{\textwidth}{!}{%
\begin{tabular}{l|ccccc}
\toprule
Simulation & IMPACC & Cluster LOCO-Split & Cluster LOCO-MP & Cluster RAMPART & Permutation \\
\midrule
Gamma ($n=200, \ p=1{,}000$)
&\textbf{ 8.8696 (0.1704)}
& 330.2923 (0.8005)
& 108.1492 (0.3053)
& 140.9914 (1.0964)
& 61.5215 (0.3026) \\
Moons and circles ($n=200, \ p=1{,}000$)
&\textbf{ 9.2214 (0.2016)}
& 354.4394 (0.7107)
& 124.4175 (0.3638)
& 224.1881 (1.2140)
& 1417.8810 (4.1919) \\
\bottomrule
\end{tabular}%
}
\end{table}

\begin{table}[htbp]
    \centering
    \resizebox{\textwidth}{!}{%
    \begin{tabular}{c|c|c|c|c|c|c|c|c|c|c}
    &
        \makecell{Cluster 0\\ \scriptsize CD4+ T reg} &
        \makecell{Cluster 1\\ \scriptsize CD4+ Naive T} &
        \makecell{Cluster 2\\ \scriptsize CD4+ Mem T} &
        \makecell{Cluster 3\\ \scriptsize CD8+ Cytotoxic T} &
        \makecell{Cluster 4\\ \scriptsize CD8+ Naive T} &
        \makecell{Cluster 5\\ \scriptsize CD14+ Monocyte} &
        \makecell{Cluster 6\\ \scriptsize CD19+ B} &
        \makecell{Cluster 7\\ \scriptsize CD34+} &
        \makecell{Cluster 8\\ \scriptsize CD56+ NK} &
        \makecell{Cluster 9 \\ \scriptsize Dendritic} \\
     \makecell{Cluster \\LOCO-MP} & 0/10 & 0/10 & 0/10 & \textcolor{blue}{5/10} & 0/10 & \textcolor{blue}{5/10} &\textcolor{blue}{4/10 }&\textcolor{blue}{3/10} & \textcolor{blue}{5/10} & 5/10 \\ 
    LRP & 0/10 & 0/10 & 0/10 & 0/10 & 0/10 & 4/10 & 3/10 & 0/10 & 4/10 & 5/10 \\        
    DEGs & 0/10 & 0/10 &\textcolor{blue}{ 1/10} & 1/10 & 0/10 & 4/10 & 3/10 & 0/10 & 4/10 &\textcolor{blue}{ 8/10} \\ 
    \end{tabular}%
    }
    \caption{Number of markers in the top-10 genes matching known markers of identified cell-type. We report in blue the highest proportion of consistent marker identified across methods for each cell types.}
    \label{tab:tab1}
\end{table}

\begin{table}[htbp]
    \centering
    \resizebox{\textwidth}{!}{%
    \begin{tabular}{c|c|c|c|c|c|c|c|c|c|c}
    &
        \makecell{Cluster 0\\ \scriptsize CD4+ T reg} &
        \makecell{Cluster 1\\ \scriptsize CD4+ Naive T} &
        \makecell{Cluster 2\\ \scriptsize CD4+ Mem T} &
        \makecell{Cluster 3\\ \scriptsize CD8+ Cytotoxic T} &
        \makecell{Cluster 4\\ \scriptsize CD8+ Naive T} &
        \makecell{Cluster 5\\ \scriptsize CD14+ Monocyte} &
        \makecell{Cluster 6\\ \scriptsize CD19+ B} &
        \makecell{Cluster 7\\ \scriptsize CD34+} &
        \makecell{Cluster 8\\ \scriptsize CD56+ NK} &
        \makecell{Cluster 9 \\ \scriptsize Dendritic} \\
     \makecell{Cluster \\LOCO-MP} & 3/10 &\textbf{ 1/10 }&\textbf{ 1/10 }& 3/10 & 3/10 & 2/10 & \textbf{3/10 }& \textcolor{red}{2/10} &\textbf{ 1/10} & \textcolor{red}{3/10} \\ 
    LRP & \textbf{1/10} & \textcolor{red}{2/10}  & \textcolor{red}{5/10} &\textbf{ 2/10} & \textbf{0/10} & \textcolor{red}{3/10} & \textcolor{red}{5/10} & \textbf{0/10} & \textcolor{red}{3/10} & \textbf{0/10 }\\        
    DEGs &  \textcolor{red}{6/10} & \textbf{1/10} & 2/10 & \textcolor{red}{4/10} & \textcolor{red}{9/10} & \textbf{1/10 }& 4/10 & 1/10 & \textcolor{red}{3/10} & 1/10\\ 
    \end{tabular}%
    }
    \caption{Number of markers in the top-10 genes corresponding to known markers for \textbf{other} cell-types (possible false positive signal). We report in red the highest proportion of inconsistent markers across methods identified for each cell type. }
    \label{tab:tab2}
\end{table}

\begin{algorithm}[htbp]
    \caption{Cluster LOCO-MP: Soft Classifiers}
    \label{alg:mp-soft}
    \begin{algorithmic}[1]
    \Require Unlabeled $\mX_1,\ldots,\mX_N\in\R^M$; minipatch sizes $(n,m)$ with $n<N$, $m<M$; number of minipatches $B$; pointwise probability error $e$.

    \State Perform minipatch consensus clustering and training:
    \begin{enumerate}[label=(\alph*),noitemsep,topsep=0pt]
        \item Randomly sample $B$ minipatches of size $n\times m$ indexed by $(I_b,F_b)_{b=1}^B$.
        \item Obtain a reference clustering $\vz_0=C_\theta(\mX)$.
        \item Cluster each minipatch to obtain $\tilde{\vz}_b=C_\theta(\mX_{I_b,F_b})$. Align $\tilde{\vz}_b$ with $\vz_{0,I_b}$ via Hungarian matching to obtain $\vz_b$, with entry $z_{b,i}$ for observation $i\in I_b$.
        \item Calculate consensus labels:
        \[ z_i^*=\arg\max_{k\in[K]} \sum_{b:\,i\in I_b}\mathbf{1}\{z_{b,i}=k\}. \]
        Break ties using the smallest class index and leave $z_i^*$ undefined when no votes are available.
        \item Train a probabilistic classifier $\hat f_b$ on $(\mX_{I_b,F_b},\vz_b)$. Express its probability vectors in reference class order, assigning zero to classes absent from the minipatch.
    \end{enumerate}

    \State Construct the Cluster LOO ensemble prediction:
    \[ \hat H_i:= \frac{\sum_{b:\,i\notin I_b}\hat f_b(\mX_{i,F_b})} {\sum_{b=1}^B\mathbf{1}\{i\notin I_b\}}. \]

    \State Construct the Cluster LOCO-LOO ensemble prediction:
    \[ \hat H_i^{-j}:= \frac{ \sum_{b:\,i\notin I_b,\;j\notin F_b} \hat f_b(\mX_{i,F_b})}{\sum_{b=1}^B\mathbf{1}\{i\notin I_b,\;j\notin F_b\}}.\]
    Leave predictions undefined when their denominator is zero.

    \State Compute Cluster LOCO-MP scores for $j=1,\ldots,M$:
    \[\hat\Delta_j:= \frac{1}{N}\sum_{i=1}^N \left[ e(z_i^*,\hat H_i^{-j}) - e(z_i^*,\hat H_i)\right],\]
    \Ensure $\{\hat\Delta_j\}_{j=1}^M$.
    \end{algorithmic}
\end{algorithm}
%\footnote{\textcolor{blue}{Cluster LOCO-LOO predictions are computed by subtracting contributions from minipatches containing feature $j$ from the LOO totals; no classifiers are refitted. For hard predictions, the implemented Hamming error separately aligns each ensemble prediction to the reference. Scalar errors and ARI use all observations; pointwise errors use the valid-observation restriction $V_j$ above.}} ensemble prediction

\begin{algorithm}[htbp]
\caption{Cluster LOCO-MP: Hard Classifiers}
\label{alg:mp-hard}
\begin{algorithmic}[1]
\Require Unlabeled $\mX_1,\ldots,\mX_N \in \R^M$; minipatch sizes $(n,m)$; number of minipatches $B$; hard-label error $\operatorname{Error}$.

\State Perform minipatch clustering and classifier training:
\begin{enumerate}[label=(\alph*),noitemsep,topsep=0pt]
    \item Randomly sample $B$ minipatches of size $n \times m$ indexed by $(I_b,F_b)_{b=1}^B$.
    \item Obtain a reference clustering $\vz_0 = C_\theta(\mX)$.
    \item Cluster each minipatch $b$ to obtain $\tilde{\vz}_b = C_\theta(\mX_{I_b,F_b})$. Align $\tilde{\vz}_b$ with $\vz_{0,I_b}$ via Hungarian matching to obtain $\vz_b$.
    \item Calculate consensus labels:
        \[ z_i^*=\arg\max_{k\in[K]} \sum_{b:\,i\in I_b}\mathbf{1}\{z_{b,i}=k\}. \]
        Break ties using the smallest class index and leave $z_i^*$ undefined when no votes are available.
    \item Train a classifier $\hat f_b$
    on $(\mX_{I_b,F_b},\vz_b)$.
\end{enumerate}

\State Construct the Cluster LOO ensemble prediction
for observation $i$:
\[\hat H(\mX_{i,:}) := \arg\max_{k\in[K]} \sum_{b:\,i\notin I_b} \mathbf{1}\{\hat f_b(\mX_{i,F_b})=k\}.\]

\State Construct the Cluster LOCO-LOO ensemble prediction
for feature $j$ and observation $i$:
\[\hat H^{-j}(\mX_{i,-j}) := \arg\max_{k\in[K]} \sum_{b:\,i\notin I_b,\;j\notin F_b} \mathbf{1}\{\hat f_b(\mX_{i,F_b})=k\}. \]

\State Compute Cluster LOCO-MP scores for $j=1,\ldots,M$:
\[\hat\Delta_j := \operatorname{Error} \!\left(\vz^*,\hat H^{-j}(\mX_{:,-j})\right)- \operatorname{Error} \!\left(\vz^*,\hat H(\mX)\right).\]

\Ensure $\{\hat\Delta_j\}_{j=1}^M$.
\end{algorithmic}

\end{algorithm}

\end{document}